\documentclass[10pt, journal]{IEEEtran}
\usepackage[ruled,linesnumbered]{algorithm2e}
\usepackage{amsmath, amssymb, amsfonts}
\usepackage{booktabs}
\usepackage{graphicx}
\usepackage{hyperref}
\usepackage{xcolor}
\usepackage[caption=false,font=footnotesize]{subfig}

\graphicspath{{figures/}}
\newtheorem{definition}{\textbf{Definition}}

\begin{document}

\title{Bridging Network Fragmentation:\\A Semantic-Augmented DRL Framework\\for UAV-aided VANETs}

\author{Gaoxiang Cao, Wenke Yuan, Huasen He, \IEEEmembership{Member, IEEE}, Yunpeng Hou, Xiaofeng Jiang, \IEEEmembership{Member, IEEE}, Shuangwu Chen, \IEEEmembership{Member, IEEE}, Jian Yang, \IEEEmembership{Senior Member, IEEE} 
\thanks{G. Cao, W. Yuan, H. He, Y. Hou, X. Jiang, S. Chen and J. Yang are with the Department of Automation, University of Science \& Technology of China.}
\thanks{H. He (hehuasen@ustc.edu.cn) and Y. Hou (hyp314@mail.ustc.edu.cn) are the corresponding authors.}
\thanks{The source code will be publicly available upon acceptance.}
}

\maketitle

\begin{abstract}
    Urban Vehicular Ad-Hoc Networks (VANETs) can become fragmented because buildings obstruct wireless links and vehicle mobility continuously changes the network topology. Unmanned Aerial Vehicles (UAVs) can serve as mobile relays, but Deep Reinforcement Learning (DRL)-based deployment often suffers from inefficient exploration because it lacks road-topology guidance. To address this problem, we propose Semantic-Augmented DRL (SA-DRL), which models network fragmentation over the road topology and aligns a pretrained Large Language Model (LLM) to generate a topology-dependent action prior from dynamic traffic states. The resulting Semantic-Augmented PPO (SA-PPO) algorithm combines this prior with the PPO policy through Logit Fusion, guiding exploration toward promising intersections while retaining adaptation through environmental returns. Simulations driven by real-world urban trajectories show that SA-PPO reaches the final converged reward of Vanilla PPO using only 28.6\% of its training episodes. It improves the average number of vehicles in connected components and the average connected-component size by 7.9\% and 8.7\%, respectively, while reducing UAV energy consumption by 21.3\%.
\end{abstract}


\section{Introduction}

\IEEEPARstart{V}{ehicular} Ad-Hoc Networks (VANETs) support both Vehicle-to-Vehicle (V2V) and Vehicle-to-Infrastructure (V2I) communication for autonomous driving in Intelligent Transportation Systems (ITS) and future sixth-generation (6G) networks \cite{6g_auto_driving}. They enable services such as cooperative sensing, traffic warnings, and computation offloading \cite{vanet_survey}. In urban canyons, however, buildings cause shadow fading \cite{shadow_fading}, while vehicle mobility changes the network topology rapidly \cite{mobility_channel}. These effects can partition a VANET into isolated subnets \cite{fragmented_subnets} and disrupt service continuity.

Roadside Units (RSUs) form the terrestrial backbone, but construction cost and site constraints limit their coverage of urban blind zones \cite{rsu_cost}. Their fixed locations also make them less responsive to sudden congestion or infrastructure disruption \cite{rsu_residence}. Unmanned Aerial Vehicles (UAVs) can instead act as mobile relays or aerial base stations \cite{uav_survey}. Their mobility and favorable probability of Line-of-Sight (LoS) links \cite{uav_los} allow them to reposition as traffic and connectivity gaps change.

Planning a UAV trajectory that bridges disconnected vehicle clusters while respecting onboard energy limits is a dynamic, non-convex optimization problem with a high-dimensional decision space \cite{np_hard}. Model-based and heuristic methods, including Mixed Integer Linear Programming (MILP), convex optimization, and potential-field methods, often require detailed environmental models and become costly in high-dimensional, time-varying settings \cite{heuristic_survey}. Deep Reinforcement Learning (DRL) offers an alternative that learns deployment decisions through interaction \cite{drl_uav}. Proximal Policy Optimization (PPO) \cite{ppo}, for example, has been applied to UAV control and trajectory planning \cite{ppo_uav}. Nevertheless, DRL exploration is usually topology-agnostic and must learn useful intersection preferences from trial and error.

Recent studies have used Large Language Models (LLMs) in UAV decision-making \cite{llm_uav, llm_uav_1}. Pretrained LLMs can process structured textual descriptions and support reasoning and planning tasks \cite{gpt4}, including action or subgoal generation for navigation \cite{llm_navigation}. This capability provides a way to encode a road graph and produce a topology-dependent action prior. Such a prior can direct DRL exploration toward relevant intersections without replacing return-based policy learning.

\subsection{Related Works}

\subsubsection{Connectivity of UAV Aided VANETs}

Prior studies have addressed UAV-assisted VANET connectivity from several perspectives, including vehicle clustering and resource optimization \cite{rel1}, drone-assisted cooperative routing \cite{rel2}, UAV-mounted Dedicated Short-Range Communications (DSRC) transceivers with Voronoi-based coverage planning \cite{rel3}, and time-varying graph models for data dissemination \cite{rel4}. Many existing approaches represent UAV deployment primarily through geometric locations, coverage regions, or local communication links, without explicitly modeling the road topology and its network-wide fragmentation structure. Such representations do not explicitly capture that vehicles move along a road graph or how fragmented vehicle clusters are connected through that graph. In addition, distance- or Signal-to-Interference-plus-Noise Ratio (SINR)-based link metrics describe local link quality rather than network-wide fragmentation. We instead construct the Road Topology Graph (RTG) and Dual Connected Graph (DCG) to quantify fragmentation and identify candidate intersections for UAV relaying.

\subsubsection{UAV Deployment Using DRL Methods}

DRL has been widely used as a model-free approach for adapting UAV movement and communication decisions in dynamic environments, including joint trajectory and task planning in obstructed scenarios \cite{rel5} and UAV placement with communication-resource control \cite{rel6}. In vehicular settings, multi-agent DRL has been combined with clustering and routing to improve reachability \cite{rel7}, while a dual-scale DRL method coordinates UAV positioning and over-the-air computation services \cite{rel8}. Although these approaches learn deployment policies from environmental interaction, they do not explicitly construct a topology-dependent action prior to guide exploration over road intersections. The agent must therefore discover useful intersections through repeated trials, which can reduce sample efficiency. Policies learned for a particular traffic distribution may also require adaptation when traffic patterns change. Our method addresses the exploration issue by aligning an LLM to produce a topology-dependent action prior and incorporating it into PPO through Logit Fusion; the PPO actor remains trainable from environmental returns.

\subsubsection{UAVs Controlled by LLM Agents}

LLMs have been used for UAV path planning from visual observations in GPS-denied environments \cite{rel9} and for estimating scene-dependent safety margins within a landing controller \cite{rel10}. Other work uses an LLM to generate reward terms for multi-UAV Q-learning \cite{rel11} or to select among RL flight policies under changing urban wind conditions \cite{rel12}. In these schemes, the LLM mainly provides high-level plans, controller parameters, reward terms, or policy selection. High-level outputs leave low-level adaptation to a separate controller, whereas LLM-generated numerical rewards can introduce unstable or biased training signals. We instead use an aligned LLM to generate a state-dependent action prior over the complete action space. Logit Fusion incorporates this prior directly into the PPO action distribution, while environmental returns continue to optimize the policy.

\subsection{Contributions}

We formulate UAV-aided VANET connectivity enhancement using the RTG and DCG. The four-stage Semantic-Augmented DRL (SA-DRL) framework then constructs action-score supervision and aligns an LLM to generate a topology-dependent action prior. Semantic-Augmented Proximal Policy Optimization (SA-PPO) incorporates this prior through Logit Fusion to guide exploration while retaining online policy adaptation. The main contributions are summarized as follows:

\begin{itemize}
    \item We propose a graph-theoretic approach to rigorously quantify the fragmentation level of VANETs in complex urban environments. By constructing the Road Topology Graph (RTG) and the Dual Connectivity Graph (DCG), we reformulate the VANET fragmentation mitigation problem as a dynamic dual graph connectivity maximization problem, which exploits the UAV to merge disjoint vehicle clusters and maximize the average size of connected components.
    \item We develop \textbf{SA-DRL}, a four-stage pipeline that  collects representative graph states, constructs action-score supervision, and aligns a pretrained LLM to generate topology-dependent action priors. These priors provide action-level guidance for subsequent reinforcement learning.
    \item We propose \textbf{SA-PPO}, which incorporates the topology-dependent action prior into PPO through \textbf{Logit Fusion} throughout online training. The prior guides action selection, while the PPO actor remains optimized using environmental returns and can correct imperfect prior preferences through continued interaction.
    \item We conduct experiments on a high-fidelity simulator driven by real-world urban trajectories. In the main comparison, SA-PPO improves two connectivity metrics by 7.9\% and 8.7\% over Vanilla PPO while reducing UAV energy consumption by 21.3\%. It also reaches the final converged reward level of Vanilla PPO using only 28.6\% of its training episodes. Tests across six traffic periods show a favorable connectivity--energy trade-off under temporal distribution shifts, and an additional Jinan experiment demonstrates that the framework retains its sample-efficiency advantage when independently trained on a different urban topology.
\end{itemize}

The remainder of this paper is organized as follows. Section \ref{sec:system_model} introduces the system model and formulates the optimization problem. Section \ref{sec:proposed_approach} elaborates on the proposed SA-DRL framework for addressing the optimization problem. Section \ref{sec:performance_evaluation} presents the experimental results and discussions. Finally, Section \ref{sec:conclusion} concludes this paper.

\section{System Model}\label{sec:system_model}

\subsection{Network Model}

\begin{figure}[ht]
    \centering
    \includegraphics[width=.6\linewidth]{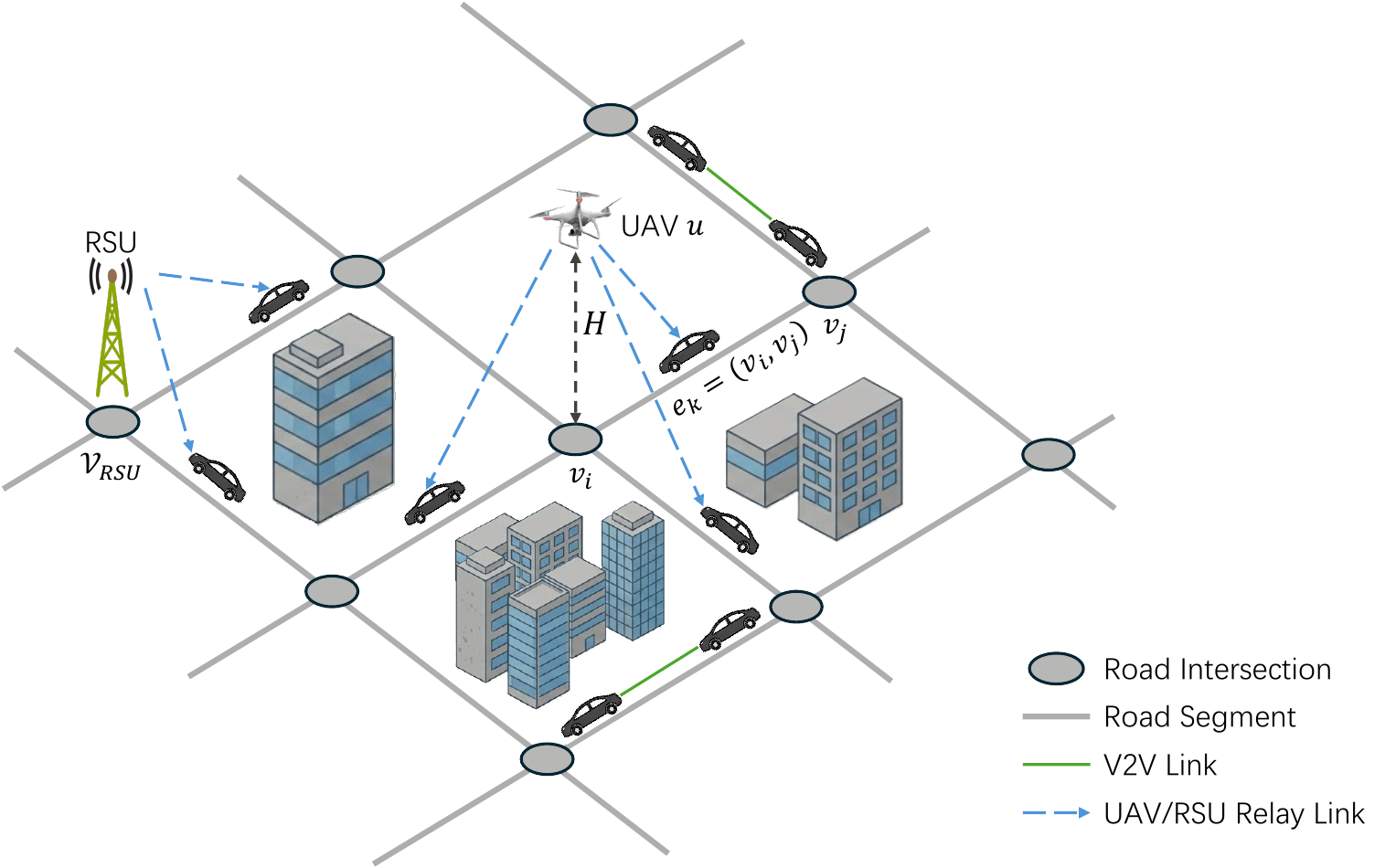}
    \caption{UAV Aided VANET Scenario.}
    \label{fig:system_model}
\end{figure}

As illustrated in Fig. \ref{fig:system_model}, we consider a UAV aided VANET system deployed within an urban region comprising $n$ intersections and $m$ road segments. The set of intersections within the task area is denoted as $\mathcal I_{road} = \{v_1, v_2, \cdots, v_n\}$, while we represent the set of road segments as $\mathcal S_{road} = \{e_1, e_2, \cdots, e_m\}$. Specifically, each intersection $v_i$ possesses fixed 2D coordinates $\boldsymbol p_i = (x_i, y_i)$, and a road segment $e_k = (v_i, v_j)$ connects intersections $v_i$ and $v_j$. Furthermore, a subset of intersections is equipped with fixed Roadside Units (RSUs), denoted as $\mathcal V_{RSU} \subset \mathcal I_{road}$. These intersections and their connecting segments maintain permanent network connectivity.

We assume that the operational timeline is discretized into a finite horizon of $T$ time slots, denoted as $\mathcal T=\{1, 2, \cdots, T\}$, where the duration of each slot is $\tau$. In the time slot $t$, there are $N(t)$ vehicles traversing the road network. If a specific vehicle $veh_l$ is located on the road segment $e_k$ at the time slot $t$, its position is denoted as $pos_{veh, l}(t) \in e_k$. While vehicle mobility follows the Intelligent Driver Model (IDM) \cite{idm}, we assume that vehicle positions are known at the beginning of each time slot. Vehicles on the road establish connections with one another via onboard wireless interfaces, forming connected clusters. However, due to high vehicular mobility, intermittent connectivity gaps persist between these clusters, resulting in VANET fragmentation. Consequently, we consider the deployment of a UAV $u$ as a communication relay to mitigate fragmentation and enhance network connectivity. The position of the UAV is denoted as $\boldsymbol p_u(t) = (x_u(t), y_u(t), H)$, where $H$ represents a fixed flight altitude. Considering that urban structures (e.g., skyscrapers) can obstruct wireless signals, we assume that the UAV's flight destination in each time slot is positioned directly above a specific intersection $v \in \mathcal I_{road}$ to maximize the probability of establishing LoS connections with ground vehicles. This assumption is adopted as a graph-level abstraction for dense urban road networks, where intersections typically provide access to multiple incident road directions and thus offer high relay efficiency for bridging fragmented vehicular subnets. Meanwhile, restricting the UAV candidate positions to intersections converts the continuous deployment problem into a finite topology-aware decision problem, which improves the computational tractability of DRL-based trajectory planning. Accordingly, the feasible UAV-position set is $\mathcal P_{road}=\{(x_i,y_i,H)\mid v_i\in\mathcal I_{road}\}$.

\subsection{Communication Model}

Given the specific characteristics of the urban environment, we adopt a probabilistic Air-to-Ground (A2G) channel model. The probability $P_{LoS}$ of establishing a LoS link between the UAV and a ground node depends primarily on the elevation angle $\theta$ \cite{los}, which is expressed as
\begin{equation}
    P_{LoS}(\theta) = \frac{1}{1 + a \cdot \exp(-b(\theta - a))}.
\end{equation}
Here, $\theta = \frac{180}{\pi} \arctan\left(\frac{H}{d}\right)$ represents the elevation angle between the UAV and the ground vehicle, $d$ is the horizontal Euclidean distance between the UAV and the ground vehicle, while $a$ and $b$ are S-curve parameters determined by the environment (e.g., dense urban, suburban, high-rise). The average path loss $L(d)$, modelled as the weighted sum of LoS and Non-Line-of-Sight (NLoS) losses, is given by
\begin{equation}
    L(d) = P_{LoS} \cdot L_{LoS}(d) + (1 - P_{LoS}) \cdot L_{NLoS}(d),
\end{equation}
where $L_{LoS}$ and $L_{NLoS}$ represent the free-space path loss plus additional attenuation loss in the two respective states, expressed as
\begin{equation}
    L_{\xi}(d) = 20\log\left(\frac{4\pi df_c}{c}\right) + \eta_{\xi}, \quad \xi \in \{LoS, NLoS\}.
\end{equation}
Here, $f_c$ is the carrier frequency, $c$ is the speed of light, while $\eta_{LoS}$ and $\eta_{NLoS}$ denote the average additional losses (shadowing loss) under LoS and NLoS conditions, respectively. Let $P_{tx}$ be the UAV transmission power, and $G$ be the sum of the receiver and transmitter antenna gains. The signal power received by a ground vehicle is given by
\begin{equation}
    P_{rx}(d) = P_{tx} + G - L(d).
\end{equation}

The maximum distance $d$ satisfying $P_{rx}(d) \geqslant P_{th}$ is defined as the UAV coverage radius $R_{cov}$, where $P_{th}$ is the minimum received power required to satisfy Quality of Service (QoS) requirements. We assume that if a UAV is positioned directly above intersection $v$, and its wireless coverage radius $R_{cov}$ exceeds half the length of the road segment, the UAV is capable of covering all vehicles situated on the road segments incident to intersection $v$. Furthermore, we assume that the communication range of the UAV under unobstructed conditions is sufficient to cover half the length of any road segment, thereby ensuring that vehicles located on the same road segment can be interconnected through multi-hop transmission. This intersection-based abstraction may be insufficient for sparse-intersection layouts or very long road segments, where a mid-segment relay position may provide higher coverage or connectivity benefits. In such cases, the road topology can be refined by inserting virtual candidate nodes along long road segments, and the proposed RTG/DCG modeling and SA-PPO framework can be directly applied to the augmented graph without changing the core algorithmic structure.

\subsection{Graph-Theoretic Model of VANET Fragmentation}

\begin{definition}[\textit{Road Topology Graph (RTG)}]
    An undirected weighted graph $\mathcal{G} = (\mathcal{V}, \mathcal{E})$ is defined as the \textit{RTG} corresponding to the urban road network $(\mathcal I_{road}, \mathcal S_{road})$ if there exist bijections $\phi: \mathcal I_{road} \to \mathcal V$ and $\psi: \mathcal S_{road} \to \mathcal E$ such that, for every road segment $s=(i_1,i_2)\in\mathcal S_{road}$ with $i_1,i_2\in\mathcal I_{road}$, $\psi(s)=(\phi(i_1),\phi(i_2))$.
    Meanwhile, the weight of a vertex $v \in \mathcal V$ is defined as
    \begin{equation}
        p_v(t) = \begin{cases} 
            0, & \text{if } \phi^{-1}(v) \text{ is currently not covered}, \\
            1, & \text{if } \phi^{-1}(v) \text{ is covered by an RSU}, \\ 
            2, & \text{if UAV } u \text{ is located at } \phi^{-1}(v).
        \end{cases}
    \end{equation}
    In particular, when the UAV $u$ is positioned at an intersection covered by an RSU, we define $p_v(t) = 2$. The weight of an edge $e \in \mathcal E$ is defined as
    \begin{equation}
        p_e(t) = \text{NumVehicles}(e,t),
    \end{equation}
    which represents the total number of vehicles located on the road segment $e$ during the time slot $t$.
\end{definition}

The vertex weights of the RTG characterize the coverage status of intersections, while the edge weights describe the traffic load on urban roads. To facilitate optimization in large-scale networks, we define \textit{c-edge} and \textit{c-vertex} as follows to characterize the connectivity of RTG vertices and edges. This allows us to abstract the complex determination of VANET connectivity into a graph-theoretic coverage problem.

\begin{definition}[\textit{Connected Vertex (c-vertex)}]
    Consider a vertex $v \in \mathcal V$, $v$ is designated as a \textit{c-vertex} in time slot $t$ if and only if at least one of the following conditions is met:
    \begin{enumerate}
        \item $p_v(t) > 0$;
        \item $\exists e \in \delta(v), p_e(t) > 0$.
    \end{enumerate}
\end{definition}

Here, $\delta(v)$ denotes the set of edges incident to vertex $v$.

\begin{definition}[\textit{Connected Edge (c-edge)}]
    Consider an edge $e = (v_1, v_2) \in \mathcal E$, $e$ is designated as a \textit{c-edge} in time slot $t$ if and only if at least one of the following conditions is met:
    \begin{enumerate}
        \item $p_e(t) > 0$;
        \item $v_1$ or $v_2$ is a c-vertex.
    \end{enumerate}
\end{definition}

Building upon the concepts of c-edge and c-vertex, we define the \textit{c-graph} as follows.

\begin{definition}[\textit{Connected Graph (c-graph)}]
    Let $\mathcal{V}_{c}(t)$ and $\mathcal{E}_{c}(t)$ denote the sets of all c-vertices and c-edges in the RTG $\mathcal{G}$ at the time slot $t$, respectively. We define the \textit{c-graph} as the subgraph $\mathcal{G}_{c}(t) = (\mathcal{V}'(t), \mathcal{E}'(t))$, where the edge set is given by $\mathcal{E}'(t) = \mathcal{E}_{c}(t)$, and the vertex set $\mathcal{V}'(t)$ is composed of $\mathcal{V}_{c}(t)$ and its associated edge endpoints, i.e.
    \begin{equation}
        \mathcal{V}'(t) = \mathcal{V}_{c}(t) \cup \{v \in \mathcal{V} \mid \exists e \in \mathcal{E}_{c}(t), v \in e\}.
    \end{equation}
\end{definition}

Since vehicles in VANETs are primarily located on road segments rather than at intersections, the connected components of $\mathcal G_c$ are insufficient to accurately quantify the degree of VANET fragmentation. This is because connected components are typically based on vertex reachability, whereas VANET connectivity should fundamentally be regarded as a form of Edge-based Connectivity. Consequently, we employ a dual graph approach by considering the following definition.

\begin{definition}[\textit{Dual Connected Graph (DCG)}]
    Consider the c-graph $\mathcal{G}_{c}(t) = (\mathcal{V}'(t), \mathcal{E}'(t))$,  for $\forall e \in \mathcal E'(t)$, based on a bijection $\varphi: \mathcal E'(t) \to \mathcal V^*(t)$, we define a unique vertex $v^* = \varphi(e) \in \mathcal V^*(t)$. For $\forall v \in \mathcal V'(t)$ and any two incident c-edges $e_1, e_2 \in \delta(v) \cap \mathcal E'(t)$, we define an edge $(\varphi(e_1), \varphi(e_2)) \in \mathcal E^*(t)$. The undirected weighted graph $\mathcal{G}^*(t) = (\mathcal{V}^*(t), \mathcal{E}^*(t))$ is termed the \textit{DCG} corresponding to $\mathcal{G}_{c}(t)$. For any $v^* \in \mathcal V^*(t)$, its weight is defined as the weight of the corresponding edge in the primal graph, which can be expressed as
    \begin{equation}
        p_{v^*}(t) = p_{\varphi^{-1}(v^*)}(t).
    \end{equation}
\end{definition}

The connected subnets (i.e., network fragments) of the VANET exhibit a one-to-one correspondence with the connected components in the DCG. Moreover, the number of vehicles within a connected subnet is equivalent to the sum of the vertex weights of the corresponding connected components. Consequently, we quantify the degree of VANET fragmentation utilizing the connected components of the DCG. Let $K(t)$ denote the number of connected components in $\mathcal G^*(t)$, with the corresponding sums of vertex weights denoted as $n_1(t), \cdots, n_{K(t)}(t)$, we define
\begin{equation}
    C(t) = \frac{1}{K(t)}\sum_{i=1}^{K(t)} n_i(t).
\end{equation}
Here, $K(t)$ represents the number of VANET network fragments, while $C(t)$ signifies the average number of vehicles within each fragment. Consequently, smaller values of $K(t)$ and larger values of $C(t)$ indicate better network connectivity.

\subsection{Energy Consumption Model}

We model the energy consumption of UAV $u$ in time slot $t$ as the sum of propulsion energy $e_f(t)$, hovering energy $e_h(t)$, and communication energy $e_c(t)$. To ensure reliable communication quality and mitigate Doppler shifts caused by UAV mobility, we adopt the hover-and-transmit paradigm \cite{uav_energy}. Similar to the approach in \cite{dtrl}, we assume a constant flight velocity for the UAV. Thus, $e_f(t)$ is proportional to the flight distance, while $e_h(t)$ and $e_c(t)$ are proportional to the hovering duration. Suppose that the UAV flies from intersection $v_i$ to intersection $v_j$ in time slot $t$ at velocity $v_u$. The total energy consumption in time slot $t$ is given by
\begin{equation}
    \label{equ:energy}
    \begin{aligned}
        E(t) =&\ e_f(t) + e_h(t) + e_c(t) \\
        =&\ (\varepsilon_1 + \varepsilon_2)\left(\tau - \frac{l(v_i, v_j)}{v_u}\right) + \varepsilon_3 \cdot \frac{l(v_i, v_j)}{v_u}. 
    \end{aligned}
\end{equation}
Here, $l(v_i, v_j)$ represents the Euclidean distance between the two intersections, while $\varepsilon_1, \varepsilon_2$, and $\varepsilon_3$ denote the power consumption for hovering, communication, and propulsion, respectively.

\subsection{Problem Formulation}

To simultaneously minimize VANET fragmentation and UAV energy consumption, we formulate the dynamic deployment problem for the UAV-assisted VANET as follows. Let $N(t)=\sum_{i=1}^{K(t)}n_i(t)$ denote the total number of vehicles at time slot $t$. Since $C(t)=N(t)/K(t)$, normalizing the connectivity term by $N(t)$ gives $C(t)/N(t)=1/K(t)$. We normalize the energy term by the maximum UAV energy consumption per time slot, denoted by $E_0$, and introduce the weighting coefficients $\alpha$ and $\beta$ to control the trade-off between connectivity and energy consumption. Both normalized terms are dimensionless. The resulting optimization problem is
\begin{align}
    & \max_{\boldsymbol P_u}\ \sum_{t\in\mathcal T}\left(\alpha\frac{C(t)}{N(t)}-\beta\frac{E(t)}{E_0}\right)
    \label{equ:opt} \\ 
    \text{s.t. } 
    &\ \boldsymbol p_u(t) \in \mathcal P_{road}, \quad \forall t \in \mathcal T,
    \tag{\ref{equ:opt}{a}} \label{equ:opt:c1} \\
    &\ l(\boldsymbol p_u(t-1), \boldsymbol p_u(t)) \leqslant v_u\tau, \quad \forall t \in \mathcal T\setminus\{1\},
    \tag{\ref{equ:opt}{b}} \label{equ:opt:c2} \\
    &\ \sum_{t\in\mathcal T}E(t) \leqslant E_{battery}.
    \tag{\ref{equ:opt}{c}} \label{equ:opt:c3}
\end{align}
where $\boldsymbol P_u = \{\boldsymbol p_u(1), \cdots, \boldsymbol p_u(T)\}$ represents the sequence of UAV positions at all time slots, and $E_{battery}$ is the UAV battery capacity. Constraint \eqref{equ:opt:c1} requires that the UAV's target destination in each time slot is an intersection. Constraint \eqref{equ:opt:c2} ensures that the flight distance in each time slot does not exceed the maximum distance feasible under the velocity limit. Constraint \eqref{equ:opt:c3} mandates that the UAV's energy is not depleted before the mission concludes.

The above problem represents a typical Mixed-Integer Non-Linear Programming (MINLP) problem. Due to the stochastic nature of the vehicular distribution $p_e(t)$ and the vastness of the state space, where the number of trajectory combinations grows exponentially with the number of intersections $n$, traditional optimization methods are computationally intractable. This challenge motivates the introduction of DRL in this study.

\section{Proposed Approach}\label{sec:proposed_approach}

\begin{figure*}[ht]
    \centering
    \includegraphics[width=.8\linewidth]{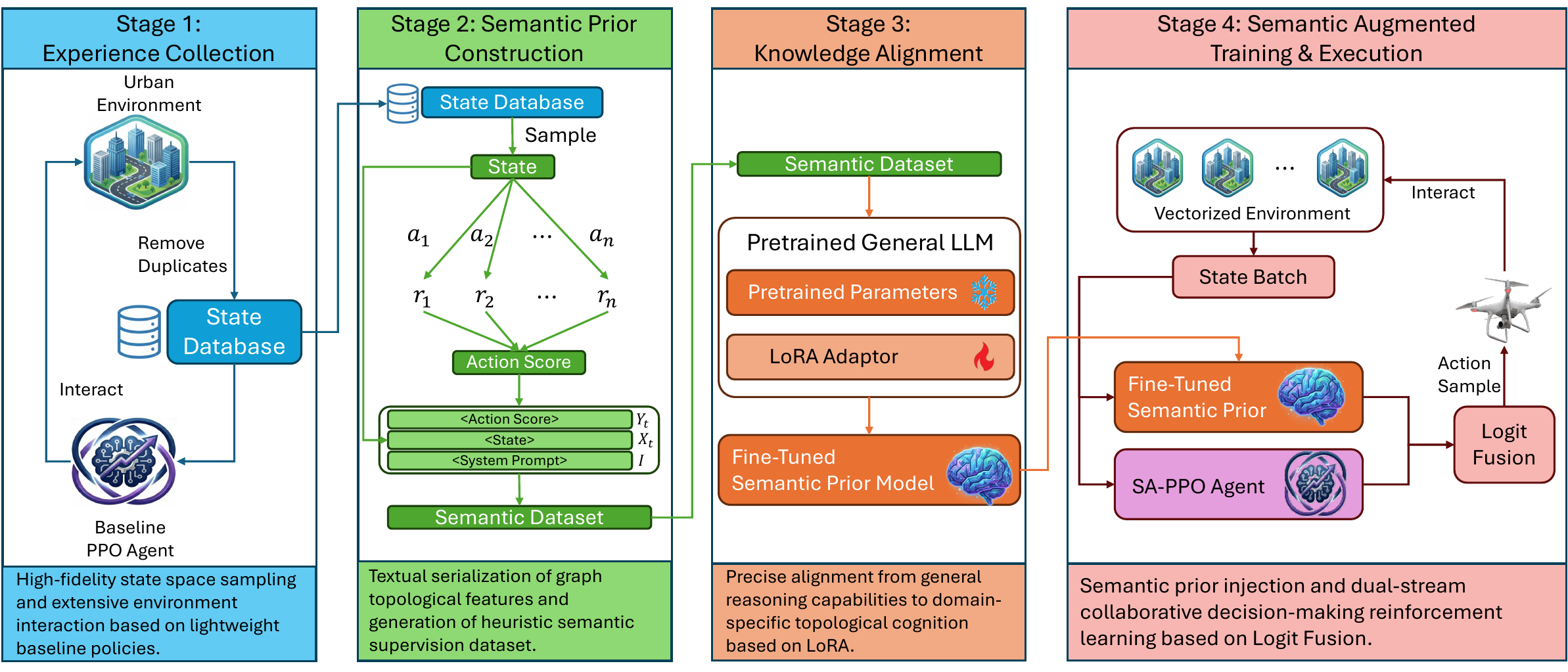}
    \caption{The four-stage pipeline of the SA-DRL framework.}
    \label{fig:system_architechture}
\end{figure*}

To address the optimization problem formulated in \eqref{equ:opt}, we propose the \textit{Semantic-Augmented Deep Reinforcement Learning} (SA-DRL) framework illustrated in Fig. \ref{fig:system_architechture}. The framework follows a four-stage pipeline. First, representative states are collected through lightweight environmental exploration. Second, these states are converted into serialized graph descriptions and paired with action-score supervision. Third, a pretrained LLM is aligned to generate topology-dependent action priors from the serialized states. Finally, we propose the \textit{Semantic-Augmented Proximal Policy Optimization} (SA-PPO) algorithm, in which the aligned prior is incorporated into PPO through Logit Fusion to guide online exploration while preserving policy adaptation through environmental interaction. This design enables the agent to leverage structured topological guidance without replacing the long-horizon reinforcement learning objective.

In the following, we first introduce the key elements of DRL. Subsequently, we elaborate on the specific details of each stage within the pipeline. Finally, we summarize the proposed SA-DRL framework.

\subsection{Key Elements of DRL}

We model the dynamic UAV deployment task as a Markov Decision Process (MDP) to solve the optimization problem formulated in \eqref{equ:opt} using DRL algorithms. An MDP is defined by the tuple $(\mathcal S,\mathcal A, r, \gamma)$, representing the state space, action space, reward function, and discount factor.

\subsubsection{Action Space}

We constrain the UAV's flight destination to always be an intersection. Therefore, $\mathcal A = \mathcal V$, and the size of the agent's action space is $n$.

\subsubsection{State Space}

We define the state at time slot $t$ as
\begin{equation}
    s_t = \{p_e(t) | e \in \mathcal E\} \cup \{p_v(t) | v \in \mathcal V\}.
\end{equation}
When used as input for the DRL algorithm, $s_t$ requires normalization.

\subsubsection{Reward Function}

We define the agent's reward at time slot $t$ as
\begin{equation}
    \label{equ:reward}
    r(t) = \alpha\cdot\frac{1}{K(t)} - \beta\cdot\frac{E(t)}{E_0},
\end{equation}
where $\alpha$ and $\beta$ are weighting coefficients, and $E_0$ represents the maximum energy consumption of the UAV per time slot.

The reward in \eqref{equ:reward} is exactly the single-slot contribution to the normalized objective in \eqref{equ:opt}. In particular, $C(t)/N(t)=1/K(t)$, while $E(t)/E_0$ is the normalized energy consumption. Thus, $\alpha$ and $\beta$ have the same meanings in the objective and reward, and summing the step-wise rewards over the mission horizon directly recovers the objective value.

\subsection{Stage 1: Experience Collection}

\begin{algorithm}[ht]
    \caption{Experience Collection}
    \label{alg:stage_1}
    \KwIn{Target number $N_{target}$, maximum number of training episodes $K_{max}$.}
    \KwOut{State database $\mathcal D_{\mathrm{state}}$.}
    Initialize empty database $\mathcal D_{\mathrm{state}} \leftarrow \varnothing$\;
    Initialize baseline PPO policy $\pi_{\theta}$ with random weights\;
    Initialize PPO rollout buffer $\mathcal{B}$\;
    \For{$k \leftarrow 1$ \KwTo $K_{max}$}{
        Reset environment and observe initial state $s_1$\;
        \For{t = 1 \rm{\textbf{to}} T}{
            Select action $a_t \sim \pi_{\theta}(\cdot | s_t)$\;
            Execute action $a_t$, observe reward $r_t$ and next state $s_{t+1}$\;
            \If{$s_t \notin \mathcal D_{\mathrm{state}}$}{
                $\mathcal D_{\mathrm{state}} \leftarrow \mathcal D_{\mathrm{state}} \cup \{s_t\}$\;
                \If{$|\mathcal D_{\mathrm{state}}| \ge N_{target}$}{
                    \Return{$\mathcal D_{\mathrm{state}}$}
                }
            }
            Store transition $(s_t, a_t, r_t, s_{t+1})$ in $\mathcal{B}$\;
            $s_t \leftarrow s_{t+1}$\;
        }
        \If{update condition met}{
            Update $\pi_{\theta}$ using data in $\mathcal{B}$ via standard PPO loss\;
            Clear $\mathcal{B}$\;
        }
    }
    \Return{$\mathcal D_{\mathrm{state}}$}
\end{algorithm}

As outlined in Algorithm \ref{alg:stage_1}, in this stage, we collect states that the agent may encounter during operation through simple environmental exploration and store them in a state database $\mathcal D_{\mathrm{state}}$. Specifically, we train a lightweight baseline PPO agent to explore the environment briefly and collect states encountered during the rollout process, which are then added to the database after deduplication. Compared to random walk sampling, the states collected by a lightweight DRL baseline are closer to the real task distribution. This approach enhances dataset quality, thereby improving training efficiency during LLM fine-tuning.

\subsection{Stage 2: Semantic Prior Construction}

\begin{algorithm}[ht]
    \caption{Semantic Prior Construction}
    \label{alg:stage_2}
    \KwIn{State database $\mathcal D_{\mathrm{state}}$, task instruction $I_{\mathrm{task}}$.}
    \KwOut{SFT dataset $\mathcal{D}_{sft}$.}
    Initialize empty dataset $\mathcal{D}_{sft} \leftarrow \varnothing$\;
    \ForEach{$s\ \rm{\textbf{in}}\ \mathcal D_{\mathrm{state}}$}{
        Initialize reward vector $\boldsymbol{r} \leftarrow \text{zeros}(n)$\;
        \ForEach{$a\ \rm{\textbf{in}}\ \mathcal{A}$}{
            Restore environment to state $s$\;
            Execute action $a$ and observe immediate reward $r_a$\;
            $\boldsymbol{r}[a] \leftarrow r_a$\;
        }
        Compute discrete action scores $Y_t$ from $\boldsymbol{r}$ using Eq. \eqref{equ:action_score}\;
        $X_t \leftarrow \text{Serialize}(s)$\;
        $D_t \leftarrow \{ \text{``instruction''}: I_{\mathrm{task}}, \text{``input''}: X_t, \text{``output''}: Y_t \}$\;
        $\mathcal{D}_{sft} \leftarrow \mathcal{D}_{sft} \cup \{D_t\}$\;
    }
    \Return{$\mathcal{D}_{sft}$}
\end{algorithm}

As shown in Algorithm \ref{alg:stage_2}, in this stage, we transform numerical graph states into semantic tokens processable by the LLM and construct a Supervised Fine-Tuning (SFT) dataset based on the state database $\mathcal D_{\mathrm{state}}$. Considering that directly inputting the graph adjacency matrix into the LLM is inefficient, we propose a concise serialization method to capture key topological features. To capture the semantic essence of VANET fragmentation, we transform the numerical state $s_t$ into a textual description of the connectivity landscape. Specifically, \textit{vehicle\_per\_edge} does not merely count vehicles but semantically represents traffic density hotspots that require coverage, while \textit{node\_weight} identifies backhaul-enabled intersections, which allows the LLM to perceive the road network not as a matrix, but as a set of disconnected vehicular clusters waiting to be bridged.

To transform the LLM into a domain expert, we construct an instruction fine-tuning dataset $\mathcal D_{sft}=\{(I_{\mathrm{task}}, X_t, Y_t)\}$. Here, $I_{\mathrm{task}}$ is the task instruction, $X_t$ is the serialized $s_t$, and $Y_t$ represents the output \textbf{action score}. Since directly computing the optimal action-value function $Q^*(s_t,\cdot)$ is intractable, we employ the immediate reward $r(s_t, \cdot)$ as an approximate action score. Although this step-wise signal does not fully capture the long-horizon return, it provides a deterministic and policy-independent measure of the immediate connectivity--energy effect of each candidate action. In the ablation study, we further compare it with a critic-bootstrapped advantage target, i.e.
\begin{equation}
    \hat A(s_t,a)=r(s_t,a)+\gamma V_{\psi}(s_{t+1}^{a})-V_{\psi}(s_t),
\end{equation}
where $V_{\psi}(\cdot)$ is estimated by the Critic of the PPO agent trained in Stage 1. The advantage-supervised variant does not improve the overall connectivity--energy and training-efficiency trade-off, since the additional value estimate introduces approximation error and dependence on the Stage 1 policy. We therefore retain the immediate reward as a low-variance step-wise proxy while explicitly acknowledging its myopic nature. A high score signifies an intersection capable of healing partitions in the DCG or merging disjoint vehicle platoons. Through SFT on this dataset, the LLM learns the mapping $f_{LLM}: X_t \to Y_t$, thereby producing topology-dependent action scores that are normalized into prior logits over the action space.

It is important to note that floating-point data not only consume more tokens but also increase the learning difficulty for the LLM. Therefore, the state vectors fed into the LLM must utilize integer components and strictly avoid normalization. Since the immediate reward defined in \eqref{equ:reward} is presented in floating-point format, we map it to the discrete interval $[0, score^*] \cap \mathbb Z$ via \eqref{equ:action_score}. In practical implementation, to maintain a compact and fixed-width representation for each action score, $score^*$ is set to 9, ensuring that each component of $Y_t$ occupies only a single character. Let $\Delta r_t=\max r(s_t,\cdot)-\min r(s_t,\cdot)$. We define
\begin{equation}
    \label{equ:action_score}
    Y_t =
    \begin{cases}
        \boldsymbol 0_{|\mathcal A|}, & \text{if } \Delta r_t=0,\\
        \operatorname{round}\!\left(\dfrac{r(s_t,\cdot)-\min r(s_t,\cdot)}{\Delta r_t}\,score^*\right),\! & \text{if } \Delta r_t>0.
    \end{cases}
\end{equation}
When all candidate actions yield the same immediate reward, there is no relative action preference available for supervision. Assigning the same zero score to all $|\mathcal A|$ actions represents an unbiased semantic prior and avoids division by zero.

\subsection{Stage 3: Knowledge Alignment}

Although prompt engineering can explicitly specify the required output format, it does not update the decision knowledge encoded in a frozen lightweight LLM. In this task, the model must generate one score for each candidate action, resulting in a score vector of length $|\mathcal{A}|$. Even with an instruction that explicitly specifies the output structure and length, the frozen base model cannot reliably produce topology-dependent action rankings. Therefore, trainable domain alignment is required to learn both the structured generation protocol and the domain-specific action-scoring function.

Among representative PEFT methods, we adopt LoRA \cite{lora} as a task-specific engineering choice. Prompt and prefix-based tuning rely on trainable virtual tokens, while bottleneck adapters introduce additional modules into Transformer layers. In contrast, LoRA learns low-rank updates to existing linear projections, which can be merged into the backbone weights during deployment. This avoids an additional inference path and is suitable for SA-PPO, where the LLM is repeatedly queried. Other PEFT methods remain viable alternatives, but their comprehensive comparison is beyond the scope of this work.

\begin{figure}[ht]
    \centering
    \includegraphics[width=.8\linewidth]{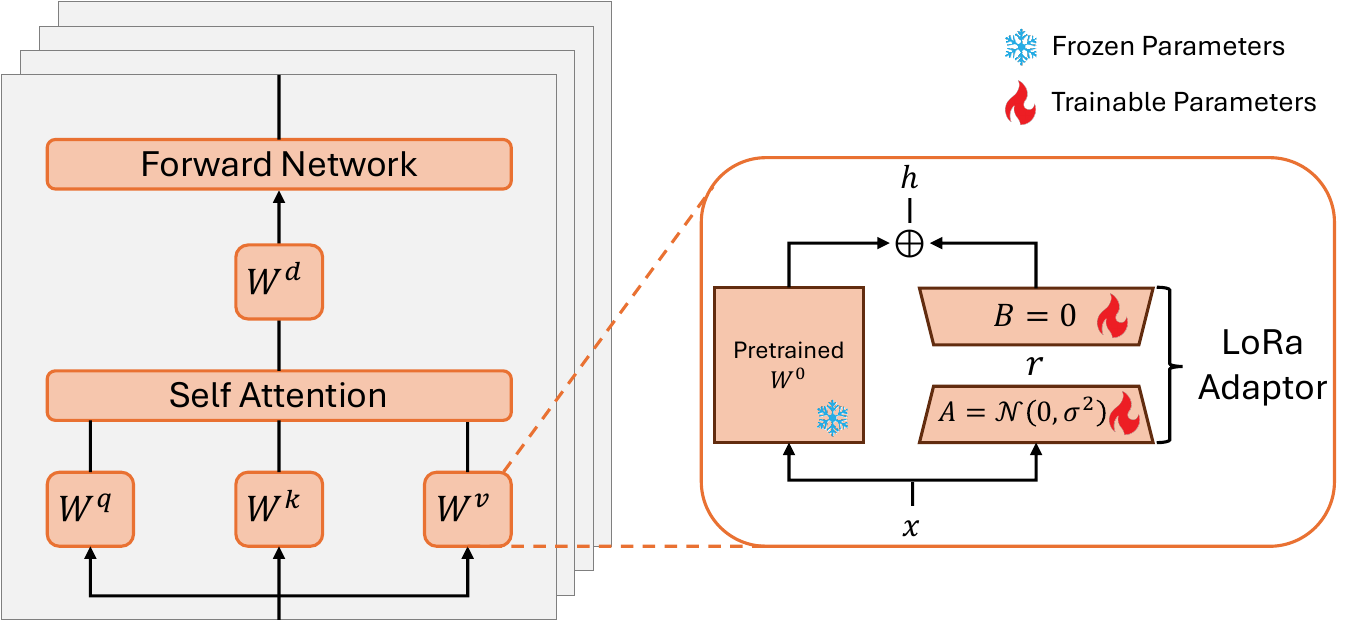}
    \caption{Schematic diagram of the LoRA fine-tuning mechanism.}
    \label{fig:lora}
\end{figure}

As shown in Fig. \ref{fig:lora}, LoRA freezes the pre-trained backbone weights $\boldsymbol{W}$ and injects trainable low-rank matrices $\boldsymbol{A}$ and $\boldsymbol{B}$ into the linear layers. The forward pass is expressed as
\begin{equation}
    \boldsymbol{Y} = \boldsymbol{X}(\boldsymbol{W}+\Delta\boldsymbol{W})
    = \boldsymbol{X}(\boldsymbol{W}+\boldsymbol{B}\boldsymbol{A}),
\end{equation}
where the rank $r$ of $\boldsymbol{A}$ and $\boldsymbol{B}$ is substantially smaller than the original model dimension. By optimizing only $\boldsymbol{A}$ and $\boldsymbol{B}$ on $\mathcal{D}_{sft}$, the model learns the topology-dependent action-scoring mapping while largely preserving the knowledge encoded in the frozen backbone.

\subsection{Stage 4: Semantic-Augmented Training \& Execution}

\begin{algorithm}[ht]
    \caption{Semantic-Augmented Training \& Execution of SA-PPO}
    \label{alg:stage_4}
    \KwIn{Fine-tuned LLM $\mathcal{M}_{SFT}$, task instruction $I_{\mathrm{task}}$, and maximum number of training episodes $K_{max}$.}
    Initialize SA-PPO agent $\mathcal{M}_{\theta}$ with random weights $\theta$\;
    Initialize rollout buffer $\mathcal{B}$\;
    \For{$k \leftarrow 1$ \KwTo $K_{max}$}{
        Reset environment\;
        \For{t = 1 \KwTo T}{
            Observe current state $s_t$\;
            $X_t \leftarrow \text{Serialize}(s_t)$\;
            $Y_t \leftarrow \mathcal{M}_{SFT}(I_{\mathrm{task}}, X_t)$; normalize $Y_t$ using Eq. \eqref{equ:semantic_logit} to obtain $\boldsymbol{z}_{LLM}$\;
            $\boldsymbol{z}_{PPO} \leftarrow \mathcal{M}_{\theta}(s_t)$\;
            Get $\tilde{\pi}(\cdot | s_t)$ with Eq. \eqref{equ:action_policy}\;
            Select action $a_t \sim \tilde{\pi}(\cdot | s_t)$\;
            Execute action $a_t$, observe reward $r_t$ and next state $s_{t+1}$\;
            Store transition $(s_t, a_t, r_t, s_{t+1}, \boldsymbol{z}_{LLM}, \boldsymbol{z}_{PPO})$ in $\mathcal{B}$\;
            $s_t \leftarrow s_{t+1}$\;
        }
        \If{update condition met}{
            Update $\mathcal M_{\theta}$ using data in $\mathcal{B}$ with the standard PPO objective\;
            Clear $\mathcal{B}$\;
        }
    }
\end{algorithm}

\begin{figure}[ht]
    \centering
    \includegraphics[width=.8\linewidth]{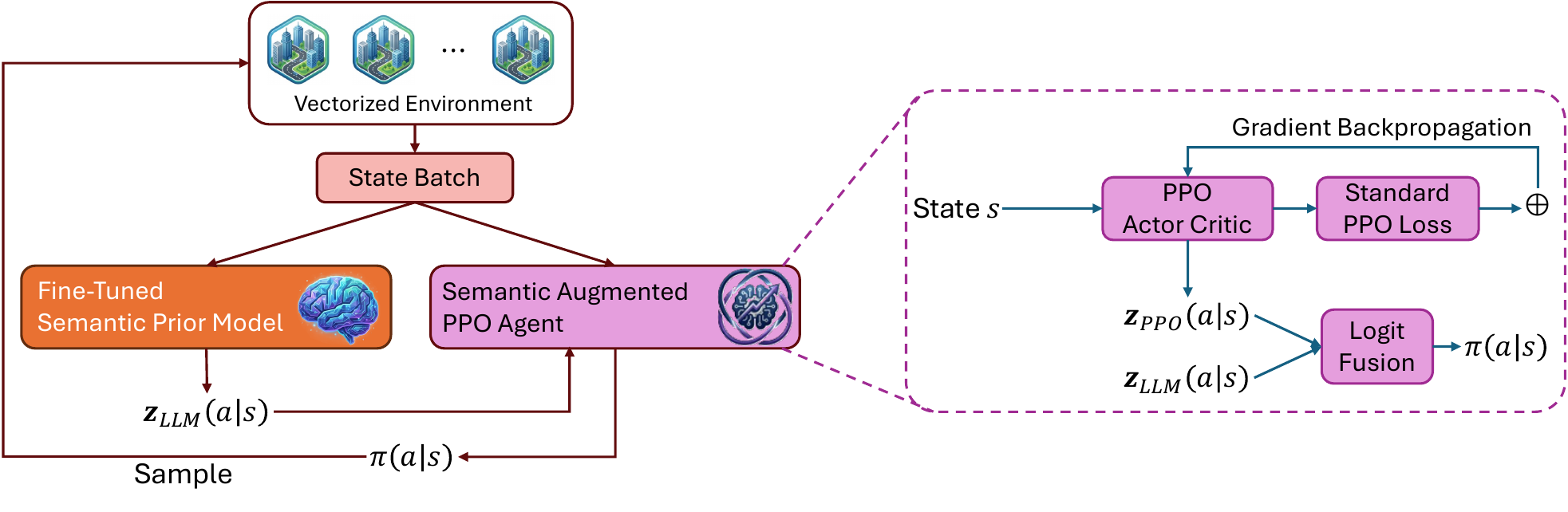}
    \caption{Schematic diagram of the SA-PPO training and execution mechanism based on Logit Fusion.}
    \label{fig:sa_ppo}
\end{figure}

In this stage, we train the final SA-PPO agent by combining the topology-dependent action prior generated by the aligned LLM with the PPO actor through Logit Fusion, as illustrated in Fig. \ref{fig:sa_ppo}. The prior provides state-dependent preferences over candidate intersections, while the PPO actor continues to learn from long-horizon environmental returns. The fused policy is used for both action sampling and PPO optimization, allowing the semantic prior to guide exploration throughout training while enabling the actor to revise imperfect or short-sighted prior preferences through continued interaction.

To combine semantic priors with real-time control policies, we design a dual-stream inference architecture. The PPO Actor network receives the state vector $s_t$ and outputs the raw logits vector $\boldsymbol{z}_{PPO}$ over the action space via a Multilayer Perceptron (MLP), i.e.
\begin{equation}
    \boldsymbol{z}_{PPO} = \text{MLP}_{actor}(s_t).
\end{equation}
Simultaneously, we serialize the current state into a text token sequence $X_t$ and obtain the semantic prior $Y_t$ output by the fine-tuned LLM. This is then normalized to serve as the raw logits vector output by the LLM, i.e.
\begin{equation}
    \label{equ:semantic_logit}
    \boldsymbol{z}_{LLM} = \frac{Y_t - \mu(Y_t)}{\sigma(Y_t) + \varepsilon}.
\end{equation}
Here, $\mu$ is the mean, $\sigma$ denotes the standard deviation, and $\varepsilon = 10^{-7}$ represents a small positive constant. For the all-zero action-score vector in \eqref{equ:action_score}, the stabilizer $\varepsilon$ in \eqref{equ:semantic_logit} yields $\boldsymbol z_{LLM}=\boldsymbol 0_{|\mathcal A|}$; Logit Fusion therefore reduces to the PPO actor policy without introducing an arbitrary action bias. The final policy $\tilde{\pi}$ is obtained by adding the normalized semantic-prior logits to the PPO actor logits and then applying the Softmax function. The semantic fusion coefficient $\lambda$ controls the relative influence of the action prior, as given by
\begin{equation}
    \label{equ:action_policy}
    \tilde{\pi}(\cdot | s_t) = \frac{\exp(\boldsymbol{z}_{PPO} + \lambda \cdot {\boldsymbol{z}}_{LLM})}{\sum_{j=1}^n \exp(\boldsymbol{z}_{PPO}^{(j)} + \lambda \cdot {\boldsymbol{z}}_{LLM}^{(j)})}.
\end{equation}
The resulting policy is used for both action sampling and PPO optimization, while the aligned LLM remains frozen. Compared with a generic policy-prior formulation, the proposed method provides an end-to-end procedure for constructing and using the prior: topology-dependent action supervision is automatically generated from dynamic graph states, the prior assigns preferences over the complete action space, and PPO can revise these preferences according to long-horizon environmental returns. Unlike reward shaping, Logit Fusion does not modify the environmental reward; unlike an imitation warm start, the action prior remains active throughout online training rather than being used only for policy initialization. Setting $\lambda=0$ recovers Vanilla PPO, whereas a larger $\lambda$ assigns greater influence to the action prior.

\subsection{High-Throughput Parallel Training System}

A major bottleneck in LLM-assisted DRL is inference latency. Considering that existing LLM inference engines can significantly increase inference speed when handling batch tasks (as opposed to serial execution), we implement a vectorized parallel training system. Instead of using a single thread for interaction sampling, we generate $N$ parallel environments to simultaneously collect a batch of states $S_{batch} = \{s_1, ..., s_N\}$. These states are serialized and queried against a shared bounded semantic-logit cache before being submitted to the LLM inference engine. Cached logits are directly reused for previously observed states, while only unique cache misses are processed by the LLM in a batch. The newly generated logits are then stored in the cache for subsequent reuse. Since the fine-tuned LLM, instruction prompt, and decoding configuration remain fixed during training, identical states deterministically produce identical semantic logits; therefore, cache reuse does not alter the resulting policy distribution. To bound CPU memory consumption, the cache contains at most $C_{\max}=100{,}000$ entries and employs a least-recently-used replacement policy. The combination of vectorized sampling, batch deduplication, batched LLM inference, and semantic-logit caching renders on-policy training utilizing LLM priors computationally feasible in large-scale urban scenarios. The semantic-logit cache can also be shared across repeated training runs with different policy-side hyperparameters, since the cached LLM outputs are independent of the fusion coefficient $\lambda$ and other PPO parameters.

The current implementation evaluates the algorithmic and computational feasibility of SA-PPO using a GPU-based inference server. In a practical deployment, the aligned LLM may be executed on an edge server or a sufficiently capable onboard computing platform. Edge-assisted inference introduces communication and queueing delays that are not explicitly modeled in the current simulator, while direct onboard execution remains constrained by the memory and power requirements of the 3B-parameter backbone. Batch inference and semantic-logit caching reduce repeated inference overhead but do not eliminate latency for previously unseen states. Therefore, the present results should be interpreted as demonstrating the feasibility of the decision framework rather than a complete real-time airborne implementation. Model quantization, prior-model distillation, and explicit communication--computation latency modeling are important directions for practical deployment.

\subsection{SA-DRL Framework}

Algorithm \ref{alg:framework} summarizes the end-to-end SA-DRL workflow by connecting the state-collection, semantic-prior construction, LLM alignment, and semantic-augmented PPO training stages described above, from the pre-trained LLM $\mathcal{M}_{base}$ and task settings to the optimized SA-PPO agent $\mathcal{M}_{\theta}$.

\begin{algorithm}[ht]
    \caption{The SA-DRL Framework}
    \label{alg:framework}
    \KwIn{Pre-trained LLM $\mathcal{M}_{base}$, task instruction $I_{\mathrm{task}}$, $K_{max}$, $N_{target}$.}
    \KwOut{Optimized SA-PPO agent $\mathcal{M}_{\theta}$.}
    Execute lightweight exploration to collect state database $\mathcal D_{\mathrm{state}}$ via \textbf{Algorithm \ref{alg:stage_1}}\;
    Serialize states and compute action scores to construct SFT dataset $\mathcal{D}_{sft}$ via \textbf{Algorithm \ref{alg:stage_2}}\;
    Initialize LoRA adapters for $\mathcal{M}_{base}$\;
    Fine-tune $\mathcal{M}_{base}$ on $\mathcal{D}_{sft}$ to obtain the domain-specific semantic model $\mathcal{M}_{SFT}$\;
    Initialize SA-PPO agent $\mathcal{M}_{\theta}$\;
    Train $\mathcal{M}_{\theta}$ online guided by semantic priors from $\mathcal{M}_{SFT}$ via \textbf{Algorithm \ref{alg:stage_4}}\;
    \Return{Converged SA-PPO agent $\mathcal{M}_{\theta}$}
\end{algorithm}

\section{Performance Evaluation}\label{sec:performance_evaluation}

\subsection{System Setup}

\begin{table}[t]
    \centering
    \caption{Simulation Environment Parameters}
    \label{tab:system_parameters}
    \begin{tabular}{clc}
        \hline
        Symbol & Parameter & Value \\
        \hline
        $\tau$ & Time-slot duration & 100 s \\
        $T_\tau$ & Total task duration & 5000 s \\
        $v_u$ & UAV flight speed & 20 m/s \\
        $\varepsilon_1$ & UAV hovering power & 100 W \\
        $\varepsilon_2$ & UAV communication power & 5 W \\
        $\varepsilon_3$ & UAV propulsion power & 200 W \\
        \hline
    \end{tabular}
\end{table}

\begin{figure}[ht]
    \centering
    \includegraphics[width=.8\linewidth]{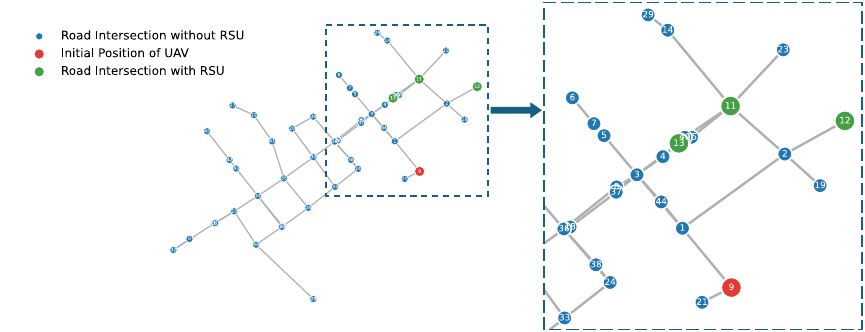}
    \caption{The RTG corresponding to the road network used in simulation.}
    \label{fig:exp_roadmap}
\end{figure}

We utilize the dataset from \cite{dataset}, which provides topological urban maps and vehicle trajectories collected by large-scale traffic surveillance systems in two Chinese cities. The original dataset contains approximately 5 million records over four days. To speed up simulations, we generate a small-scale subset based on the data for a specific area in Shenzhen on April 16, 2021, consisting of 47 nodes, 88 edges, and 5,000 trajectory records. Based on this data set, we construct a Python simulation system to implement the four-stage SA-DRL pipeline. Figure \ref{fig:exp_roadmap} depicts the RTG of the simulated network, where the red node marks the UAV's starting position (Node 9), green nodes denote intersections equipped with RSUs (Nodes 11-13), and blue nodes represent standard road intersections. Following this setup, we conduct comparative experiments and ablation studies to verify the algorithm's effectiveness. In terms of implementation, we use PyTorch for the SA-PPO algorithm, LLaMA Factory \cite{llama_factory} for LLM fine-tuning, and the vLLM \cite{vllm} framework for inference acceleration. The experimental hardware consists of an Intel Core i7-14700KF @ 3.42 GHz CPU and an NVIDIA GeForce RTX 4080 Super GPU. The principal simulation and UAV energy-model parameters are summarized in Table \ref{tab:system_parameters}, where the UAV energy parameters are selected within representative ranges reported in existing UAV-assisted wireless communication studies \cite{hover_parameter, communication_parameter, propulsion_parameter}.

\subsection{Analysis of the Semantic Prior Model}
\label{sec:semantic_prior_model}

\subsubsection{Performance Evaluation of Semantic Knowledge Alignment}

In addition to standard LLM evaluation metrics such as the Bilingual Evaluation Understudy (Cumulative 4-gram score, BLEU-4) and the Recall-Oriented Understudy for Gisting Evaluation (based on Longest Common Subsequence, ROUGE-L), we introduce three specific evaluation metrics tailored to the algorithm's characteristics:

\begin{itemize}
    \item \textbf{JSON Parsing Success Rate} (JSON PSR): Given that the DRL training loop requires automated interaction, the LLM output must strictly adhere to the defined JSON schema. We define the JSON Parsing Success Rate as the proportion of generated responses that can be successfully decoded by a standard JSON parser without syntax errors (e.g., missing brackets, unescaped characters). We have
    \begin{equation}
        PSR = \frac{1}{N} \sum_{i=1}^{N} \mathbb{I}(\text{isValidJSON}(o_i)).
    \end{equation}
    \item \textbf{Kendall's Rank Correlation Coefficient} (Kendall's $\tau$): This metric quantifies the ordinal association between the semantic scores predicted by the LLM and the ground-truth heuristic rewards. Unlike absolute value-based metrics (e.g., Mean Squared Error), Kendall's $\tau$ focuses on the relative ranking of candidates, which is crucial for DRL agents that select actions based on relative dominance. Given two sequences of equal length $\mathbf{x}, \mathbf{y}$, we enumerate all pairs of positions $(i, j)$. When $(x_i - x_j)(y_i - y_j) > 0$ (concordant pairs), we let the count be $C$. As for $(x_i - x_j)(y_i - y_j) < 0$ (discordant pairs), we let the count be $D$. Then, Kendall's $\tau$ is defined as
    \begin{equation}
        \tau = \frac{C - D}{C + D}.
    \end{equation}
    A larger $\tau$ indicates a higher correlation in ranking between the two sequences. $\tau = 1$ implies identical ranking, while $\tau = -1$ implies completely opposite ranking.
    \item \textbf{Top-$k$ Hit Rate} ($HR_k$): Beyond global ranking correlation, a critical metric is ensuring that optimal actions are included within the high-probability candidate set. We adopt a set-based Top-$k$ Hit Rate. Specifically, for each state, we sort both the predicted semantic scores and the ground-truth heuristic rewards in descending order to extract two index subsets: the top-$k$ predicted set $\mathcal{S}_{pred}^{(k)}$ and the top-$k$ ground truth set $\mathcal{S}_{gt}^{(k)}$. To calculate the ratio of the cardinality of their intersection, we define
    \begin{equation}
        HR_k = \frac{1}{M} \sum_{i=1}^{M} \frac{|\mathcal{S}_{pred, i}^{(k)} \cap \mathcal{S}_{gt, i}^{(k)}|}{k},
    \end{equation}
    where $M$ is the number of samples. This metric quantifies the overlap ratio, indicating the accuracy with which the LLM captures the set of dominant topological nodes. In the subsequent presentation of results, we set $k = 10$.
\end{itemize}

To evaluate the impact of different backbone architectures on topological reasoning and instruction-following capabilities, we benchmark four state-of-the-art lightweight LLMs: Qwen2.5-3B, Qwen3-4B, Llama3.2-3B, and Gemma3-4B. We focus on models with fewer parameters to ensure the feasibility of deployment on resource-constrained UAVs. The comparative results are summarized in Table \ref{tab:fine_tune}.

\begin{table}[ht]
    \centering
    \caption{Performance Comparison of Prompt-Only and LoRA-Based Knowledge Alignment. Kendall's $\tau$, JSON PSR, and $HR_{10}$ are reported as percentages.}
    \label{tab:fine_tune}
    \setlength{\tabcolsep}{4pt}
    \begin{tabular}{lccccc}
        \hline
        LLM & Qwen2.5 & Qwen2.5 & Qwen3 & Llama3.2 & Gemma3 \\
        Parameters & 3B & 3B & 4B & 3B & 4B \\
        \textbf{Alignment} & Prompt-only & LoRA & LoRA & LoRA & LoRA \\
        \hline
        BLEU-4 & 0.79 & \textbf{75.10} & \underline{70.32} & 44.64 & 55.50 \\
        ROUGE-L & 18.55 & \textbf{81.19} & \underline{78.42} & 69.81 & 74.84 \\
        Kendall's $\tau$ & -- & \underline{86.14} & 78.82 & 85.60 & \textbf{88.89} \\
        JSON PSR & 0.00 & \textbf{100.00} & \textbf{100.00} & \textbf{100.00} & \textbf{100.00} \\
        $HR_{10}$ & 0.00 & 55.02 & 51.11 & \underline{55.53} & \textbf{58.60} \\
        \hline
    \end{tabular}
\end{table}

As demonstrated by the first two columns of Table \ref{tab:fine_tune}, prompt-only alignment remains insufficient for the proposed action-scoring task. In the considered experimental topology, the action space contains 47 candidate intersections; therefore, the prompt-only Qwen2.5-3B model is explicitly instructed to return exactly 47 integer scores in a single JSON array without any additional text. Nevertheless, it achieves a JSON PSR of 0.00\%, a BLEU-4 score of 0.79, a ROUGE-L score of 18.55, and an $HR_{10}$ of 0.00\%. Its outputs frequently contain explanatory text or incomplete arrays. Even when numerical values can be extracted through post-processing, the predicted scores often degenerate into monotonic sequences or repeated values and thus cannot provide a meaningful action-ranking prior. This indicates that output constraints and post-processing alone cannot enable the frozen lightweight model to learn the domain-specific mapping from VANET states to action preferences.

In contrast, all four LoRA-aligned models achieve a JSON PSR of 100.00\%, demonstrating reliable compliance with the required output protocol. Qwen2.5-3B achieves the highest BLEU-4 and ROUGE-L scores of 75.10 and 81.19, respectively, together with a Kendall's $\tau$ of 86.14\% and an $HR_{10}$ of 55.02\%. Qwen3-4B achieves competitive textual similarity but relatively lower ranking performance, with a Kendall's $\tau$ of 78.82\% and an $HR_{10}$ of 51.11\%. Llama3.2-3B achieves a Kendall's $\tau$ of 85.60\% and the second-highest $HR_{10}$ of 55.53\%. Gemma3-4B provides the strongest ranking performance, achieving the highest Kendall's $\tau$ of 88.89\% and $HR_{10}$ of 58.60\%. Since semantic accuracy alone is insufficient to determine the most suitable backbone for the high-frequency interaction required by SA-PPO, we further compare their inference efficiency in the following subsection.

\subsubsection{Inference Efficiency Analysis}

\begin{figure}[ht]
    \centering
    \includegraphics[width=.8\linewidth]{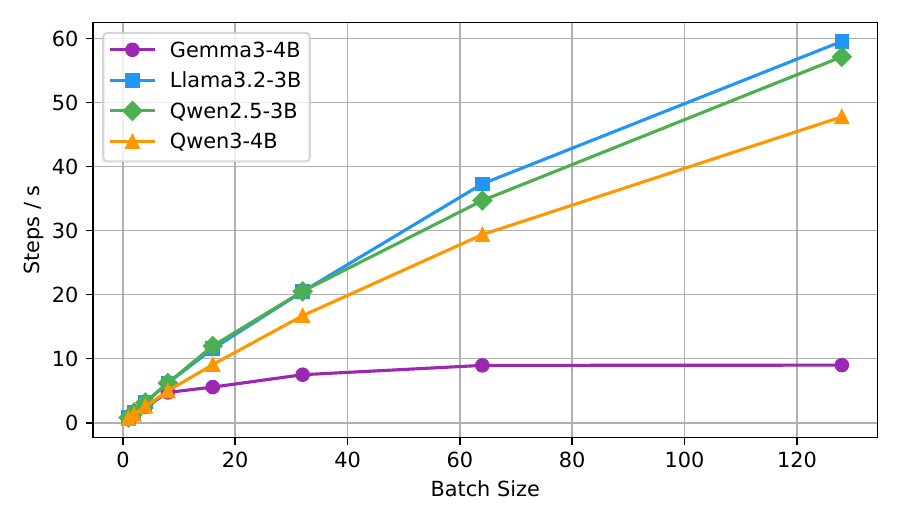}
    \caption{Inference efficiency comparison for different LLM architectures as batch size varies.}
    \label{fig:infer_efficiency}
\end{figure}

While semantic alignment is imperative, the inherent high-frequency interaction characteristic of DRL imposes stringent constraints on inference latency. Although our proposed parallel training system enables batch inference, the scalability of the underlying LLM remains a critical bottleneck determining the overall training efficiency. As illustrated in Fig. \ref{fig:infer_efficiency}, we observe a critical performance divergence as the batch size increases. The throughput of Gemma3-4B (purple line) saturates rapidly, stagnating at approximately 9 steps/s regardless of the increase in batch size. This indicates a severe VRAM bottleneck (e.g. the VRAM available for the KV-Cache is insufficient), likely attributable to its larger parameter count and architectural overhead, rendering it unsuitable for parallelized training environments. In contrast, Qwen2.5-3B (green line) demonstrates superior scalability. Its throughput exhibits a linear increase with batch size, peaking at approximately 58 steps/s at a batch size of 128, which is more than 6 times faster than Gemma3-4B. This high-throughput capability enables the SA-PPO agent to collect samples and update its policy significantly faster, thereby substantially reducing the total training duration. In summary, although Gemma3-4B holds a marginal advantage in semantic reasoning, its suboptimal inference efficiency imposes an unacceptable computational burden on the system. Consequently, we select Qwen2.5-3B as the backbone network and set the batch size to 128, thereby striking an optimal balance between reasoning capability and system efficiency.

\subsubsection{Visual Analysis of Semantic Alignment}

\begin{figure}[ht]
    \centering
    \includegraphics[width=.8\linewidth]{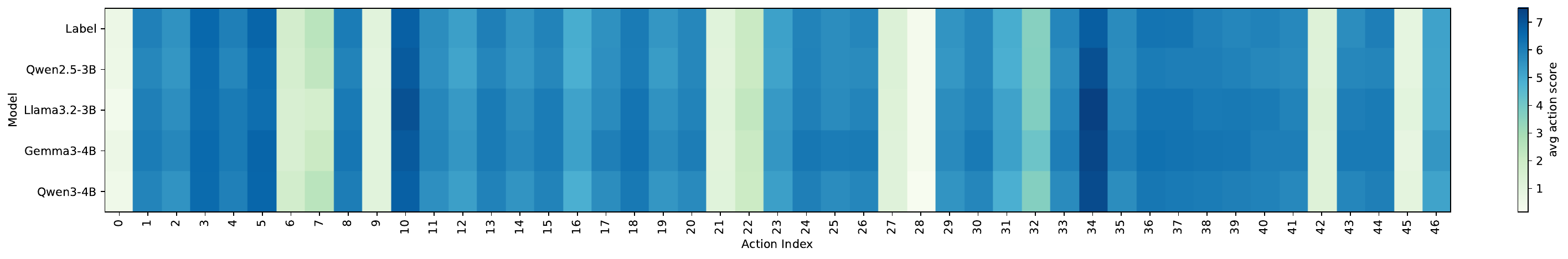}
    \caption{Heatmap comparison of action scores predicted by different fine-tuned LLMs against ground truth.}
    \label{fig:action_score}
\end{figure}

To provide a more intuitive understanding of the efficacy of semantic alignment, we visualize the action score distributions generated by the LLMs. Fig. \ref{fig:action_score} presents a heatmap comparison between the average semantic scores predicted by different fine-tuned models and the ground truth (Label) on the test set. In this visualization, the x-axis represents the index of intersection nodes (Action Index), while the color intensity corresponds to the average action score. Variations in color distribution reflect the differing learning capabilities of the models regarding the environment. However, the overall trends reveal the intrinsic topological characteristics of the urban landscape. Since the heatmap represents aggregated scores across all test states, the vertical bands consistently exhibiting high scores (e.g., Nodes 34 and 10) identify nodes possessing high global centrality or strategic importance. By cross-referencing with the road topology map in Fig. \ref{fig:exp_roadmap}, we observe that these high-scoring indices correspond to critical traffic hubs (notably, they are all cut vertices of the graph). This correspondence confirms that the fine-tuned LLMs have successfully internalized the spatial structure of the road network, learning to prioritize topologically critical nodes while abstracting away transient traffic fluctuations.

\begin{figure}[ht]
    \centering
    \includegraphics[width=.8\linewidth]{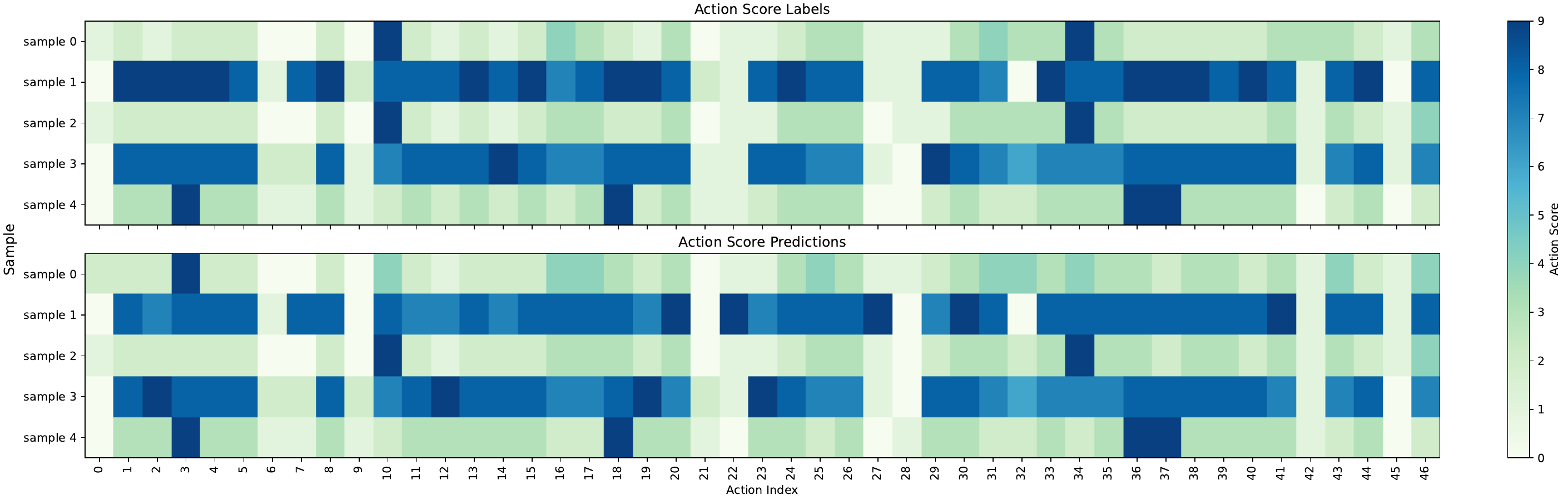}
    \caption{Comparison of action score predictions by Qwen2.5-3B against ground truth.}
    \label{fig:action_score_qwen2.5}
\end{figure}

Furthermore, to validate the stability of the selected backbone model, Fig. \ref{fig:action_score_qwen2.5} details the prediction performance of Qwen2.5-3B across multiple random samples. The upper row displays the ground truth labels, while the lower row shows the model predictions. The results indicate that the model successfully generalizes across diverse topological states, consistently identifying key intersections and reproducing the sparse distribution characteristics of the reward function. This confirms that the LoRA-based fine-tuning has successfully aligned the reasoning capabilities of the LLM with the domain-specific requirements of urban vehicular networks.

\subsection{Performance Evaluation of SA-PPO}

\subsubsection{Comparative Experiments}

To validate the effectiveness of the proposed SA-PPO algorithm in dynamic UAV deployment, we compare it against three representative DRL baselines. These baselines encompass standard On-Policy and Off-Policy algorithms, as well as graph-reinforced DRL methods.

\begin{itemize}
    \item \textbf{Soft Actor-Critic} (SAC): An Off-Policy DRL algorithm utilized in \cite{cmp_algo} to optimize the Age of Information (AoI) in UAV-aided vehicular edge computing. SAC enhances exploration by simultaneously maximizing reward and policy entropy.
    \item \textbf{Vanilla PPO} \cite{ppo}: The standard Proximal Policy Optimization algorithm (On-Policy) without any semantic or structural enhancements. The agent observes raw state vectors and learns solely through trial-and-error interactions with the environment.  
    \item \textbf{GAT-PPO}: A PPO variant that employs a Graph Attention Network (GAT) as the feature extractor, replacing the standard Multi-Layer Perceptron (MLP). Since our environment is modeled based on RTG, GAT-PPO is designed to explicitly extract structured data from the graph.
    \item \textbf{SA-PPO} (Ours): The proposed algorithm which integrates the domain-specific topological expertise of a fine-tuned LLM into the PPO training loop via the Logit Fusion mechanism.
\end{itemize}

\begin{figure}[ht]
    \centering
    \includegraphics[width=.8\linewidth]{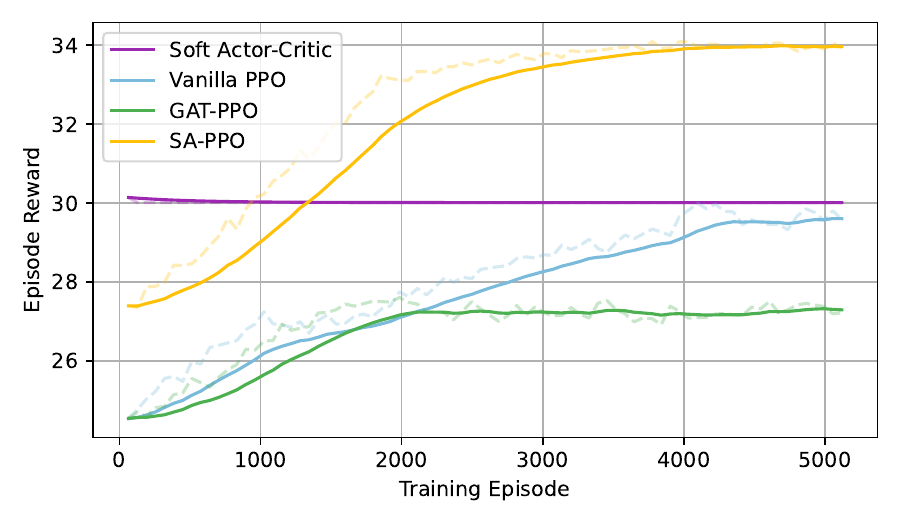}
    \caption{Training reward curves for different algorithms.}
    \label{fig:train_reward}
\end{figure}

\begin{figure}[ht]
    \centering
    \includegraphics[width=.8\linewidth]{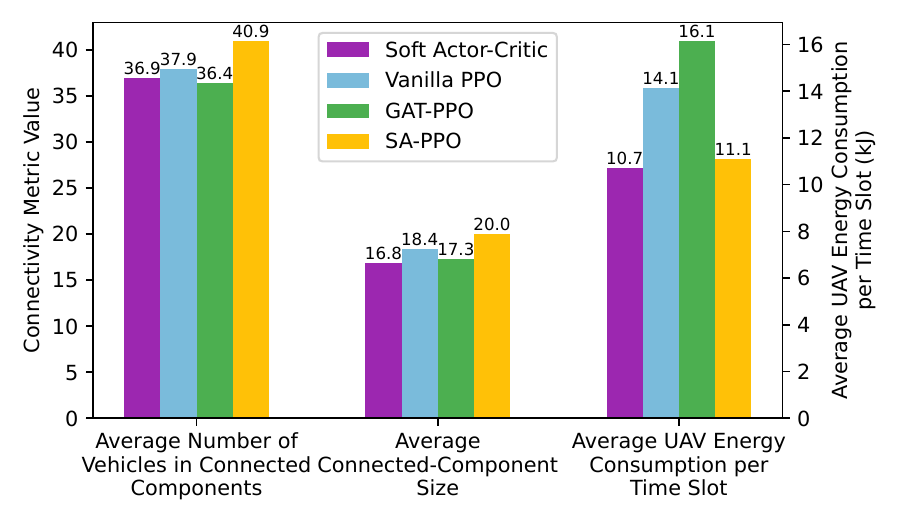}
    \caption{Comparison of VANET connectivity and UAV energy consumption metrics.}
    \label{fig:algo_cmp}
\end{figure}

As shown in Fig. \ref{fig:train_reward}, SA-PPO demonstrates substantially higher online RL sample efficiency than the baseline algorithms. To make this comparison verifiable, we define the ``baseline performance level'' as the final converged reward of Vanilla PPO, approximately 29.5 in Fig. \ref{fig:train_reward}, because Vanilla PPO is the direct counterpart of SA-PPO without semantic augmentation. Based on the first crossing points of the smoothed curves, SA-PPO reaches this level after approximately 1,200 episodes, whereas Vanilla PPO reaches it after approximately 4,200 episodes. Therefore, SA-PPO uses only $1200/4200 \times 100\% = 28.6\%$ as many online RL episodes, corresponding to a 71.4\% reduction in online training interactions. This comparison does not represent end-to-end computational efficiency: state collection and label construction require less than one minute, LoRA fine-tuning requires approximately 20 minutes, and Stage-4 SA-PPO training with an initially empty cache requires approximately 7 minutes, compared with approximately 1.5 minutes for Vanilla PPO and 30 minutes for SAC. Thus, SA-PPO exchanges additional computation and model capacity for fewer online interactions and stronger final performance rather than reducing total one-shot training cost.

The quantitative metrics in Fig. \ref{fig:algo_cmp} further characterize the connectivity--energy trade-off. For the two connectivity metrics, SA-PPO obtains an average of 40.9 vehicles in connected components and an average connected-component size of 20.0, compared with 37.9 and 18.4 for Vanilla PPO. These results correspond to relative improvements of $(40.9-37.9)/37.9 \times 100\% = 7.9\%$ and $(20.0-18.4)/18.4 \times 100\% = 8.7\%$, respectively. Meanwhile, the UAV energy consumption computed by Eq. \eqref{equ:energy} is reduced from 14.1~kJ for Vanilla PPO to 11.1~kJ for SA-PPO, yielding a 21.3\% reduction. Although SAC has slightly lower energy consumption, it achieves much weaker connectivity, indicating a passive low-mobility policy.

\begin{figure}[ht]
    \centering
    \subfloat[Soft Actor-Critic]{
        \includegraphics[width=0.4\linewidth]{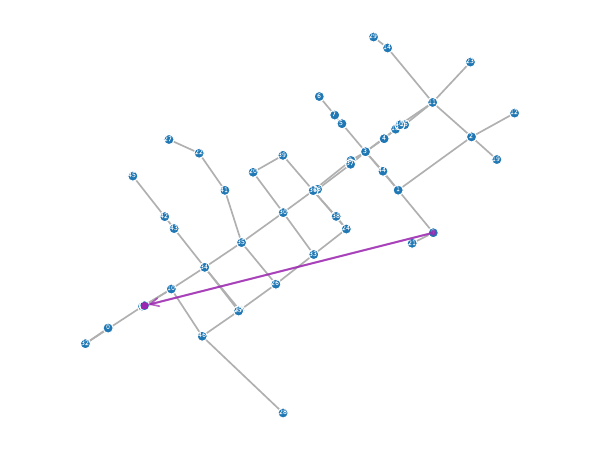} 
    }
    \subfloat[Vanilla PPO]{
        \includegraphics[width=0.4\linewidth]{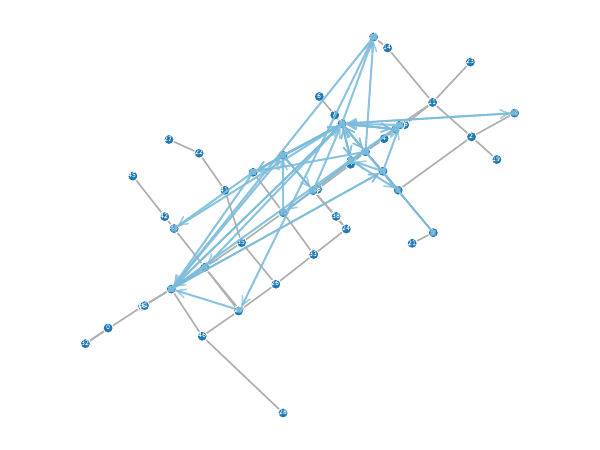}
    }
    \\
    \subfloat[GAT-PPO]{
        \includegraphics[width=0.4\linewidth]{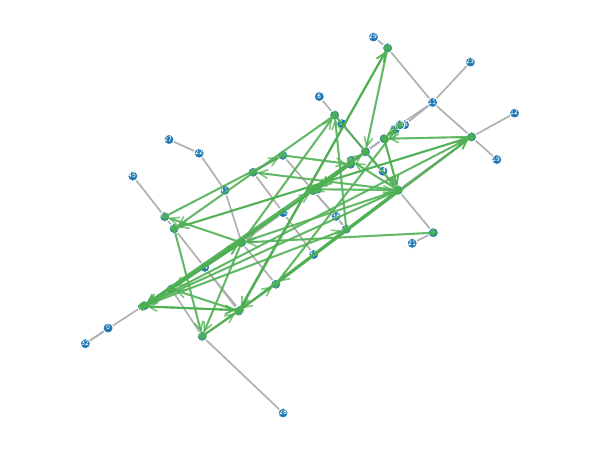}
    }
    \subfloat[SA-PPO]{
        \includegraphics[width=0.4\linewidth]{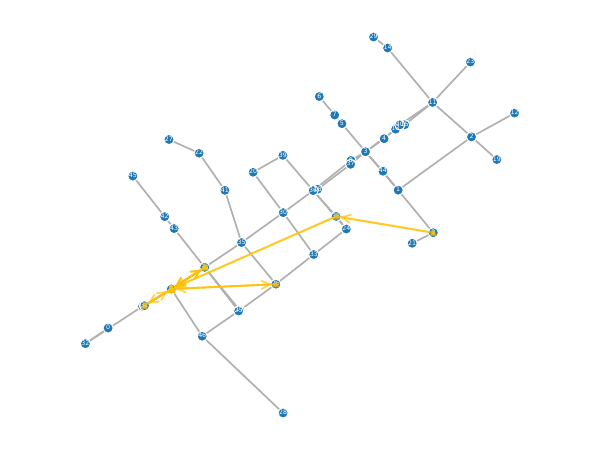}
    }
    \caption{UAV trajectory visualization under different algorithms.}
    \label{fig:trajectory}
\end{figure}

As shown in Fig. \ref{fig:train_reward} and Fig. \ref{fig:algo_cmp}, GAT-PPO does not provide a consistent advantage over Vanilla PPO despite employing a Graph Neural Network (GNN). Its average number of vehicles in connected components is lower (36.4 versus 37.9), and its average connected-component size is also lower (17.3 versus 18.4). Its average UAV energy consumption is also higher, at approximately 16.1~kJ per time slot, compared with 14.1~kJ for Vanilla PPO and 11.1~kJ for SA-PPO. The trajectory in Fig. \ref{fig:trajectory} suggests frequent movements between adjacent nodes in response to local traffic fluctuations, which is consistent with the higher energy consumption. In this setting, increasing the feature extractor's complexity alone therefore does not yield a better connectivity--energy trade-off.

Furthermore, to qualitatively examine the learned behaviors, we visualize the UAV trajectories generated by different algorithms in Fig. \ref{fig:trajectory}. SA-PPO moves toward a small set of central intersections and then remains comparatively stationary, which is consistent with its average UAV energy consumption of approximately 11.1~kJ per time slot. Vanilla PPO and GAT-PPO exhibit more extensive movement, whereas SAC remains nearly static and consequently attains low energy consumption but weak connectivity. These trajectories provide qualitative support for the distinct connectivity--energy trade-offs reported in Fig. \ref{fig:algo_cmp}. They are not used as independent evidence of optimality.

\subsubsection{Ablation Studies}

To further examine the contribution of semantic priors and the effectiveness of the Logit Fusion mechanism, we conduct ablation studies by comparing the proposed SA-PPO with four variants:

\begin{enumerate}
    \item \textbf{w/o LLM (Vanilla PPO)}: The semantic branch is removed, and the agent learns solely through the PPO actor. This serves as a benchmark without semantic guidance.
    
    \item \textbf{w/o LLM w/ MLP (MLP-guided PPO)}: The LLM is replaced by a lightweight MLP trained using the same state representation and immediate-reward-based action-score labels. Its output is normalized and incorporated into PPO using the same Logit Fusion mechanism and fusion coefficient. This matched variant evaluates the effect of changing the prior-model backbone while keeping the supervision and integration settings unchanged.
    
    \item \textbf{w/ Advantage SFT}: The immediate-reward-based supervision is replaced by advantage-based labels during LLM fine-tuning. Specifically, for each collected state $s_t$, the environment is restored and each candidate action $a$ is executed once to obtain the immediate reward $r(s_t,a)$ and the corresponding next state $s_{t+1}^{a}$. The action label is then computed as $\hat{A}(s_t,a)=r(s_t,a)+\gamma V_{\psi}(s_{t+1}^{a})-V_{\psi}(s_t),$ where $V_{\psi}(\cdot)$ is the value function estimated by the Critic of the PPO agent trained in Stage 1. The resulting advantage values are discretized into action scores for SFT. This variant evaluates whether incorporating the Critic-estimated future return into the semantic labels improves the final policy.
    
    \item \textbf{w/o Logit Fusion (Pure Semantic Policy)}: The RL branch is removed, and actions are selected by directly sampling from the fine-tuned LLM output distribution ($a \sim \text{Softmax}(z_{LLM})$). This variant is used to evaluate the quality of the raw semantic priors without RL adaptation.
\end{enumerate}

\begin{figure}[ht]
    \centering
    \subfloat[Train curves of SA-PPO and its variants.]{
        \label{fig:ablation_reward}
        \includegraphics[width=0.8\linewidth]{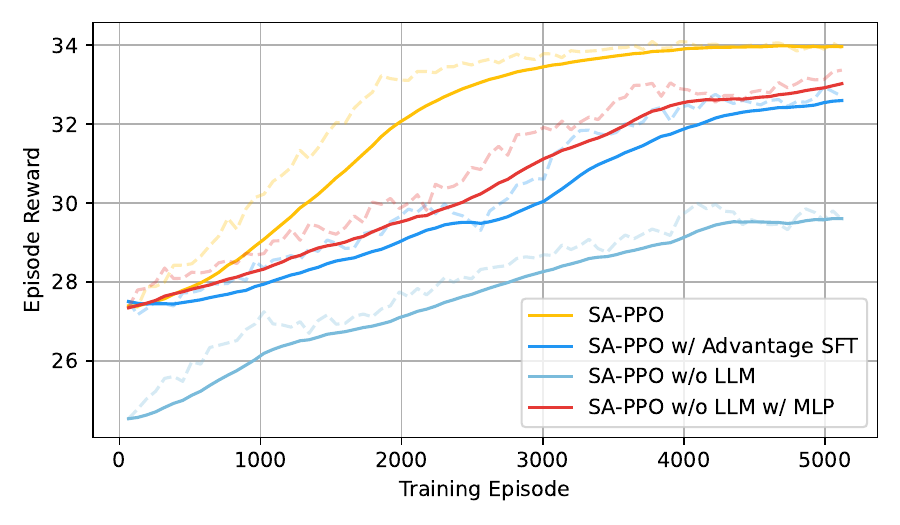} 
    }
    \\
    \subfloat[VANET connectivity and UAV energy consumption metrics of SA-PPO and its variants.]{
        \label{fig:ablation_algo_cmp}
        \includegraphics[width=0.8\linewidth]{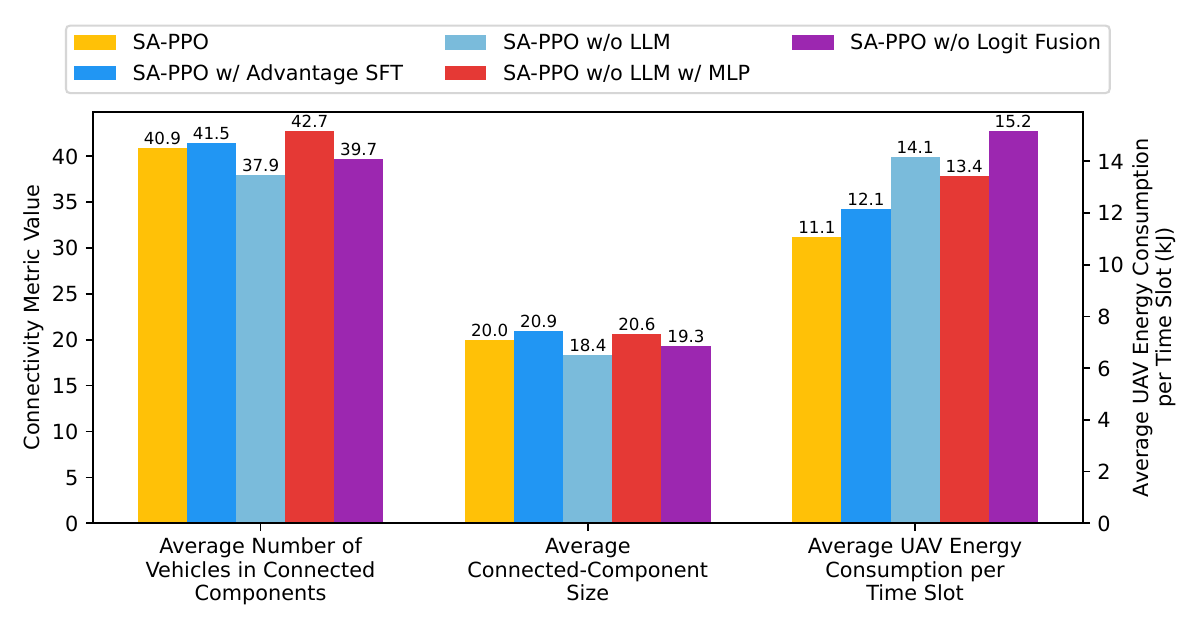}
    }
    \caption{Ablation study results comparing SA-PPO with its variants.}
    \label{fig:ablation_study}
\end{figure}

As shown in Fig. \ref{fig:ablation_reward}, SA-PPO, MLP-guided PPO, and SA-PPO w/ Advantage SFT start from similar episode-reward levels. This is expected because the three variants use the same PPO architecture and initialization, while their prior outputs are normalized and incorporated using the same Logit Fusion coefficient. Moreover, SA-PPO and MLP-guided PPO are trained using the same immediate-reward-based state--action supervision, leading to comparable initial action preferences before substantial online policy adaptation. As training proceeds, however, their performance gradually diverges. SA-PPO enters the high-reward region earlier and ultimately achieves the highest episode reward, whereas MLP-guided PPO improves more slowly but still clearly outperforms Vanilla PPO. This result confirms that the automatically generated state--action supervision itself provides useful exploration guidance, while the choice of prior model further affects the quality of the trajectories collected during online learning.

The Advantage-SFT variant also begins from a similar reward level but converges more slowly and reaches a lower final reward than SA-PPO. Its supervision includes the Critic-estimated future-value term $\gamma V_{\psi}(s_{t+1}^{a})-V_{\psi}(s_t)$, which introduces approximation error and dependence on the Stage 1 policy into the action ranking. Since the resulting prior remains fixed during subsequent SA-PPO training, errors in the estimated action preferences can affect the trajectories collected by PPO and become progressively reflected in later policy updates. In the considered setting, the additional future-value estimate therefore does not provide a more effective semantic prior than the deterministic immediate-reward supervision.

The detailed connectivity and energy metrics are presented in Fig. \ref{fig:ablation_algo_cmp}. MLP-guided PPO achieves the highest average number of vehicles in connected components, at 42.7, while the Advantage-SFT variant obtains the largest average connected-component size, at 20.9. However, both variants consume more UAV energy than SA-PPO, requiring 13.4~kJ and 12.1~kJ, respectively, compared with 11.1~kJ for SA-PPO. The pure semantic policy also outperforms Vanilla PPO in both connectivity metrics, indicating that the aligned prior model captures useful action preferences even without online policy adaptation. Nevertheless, it incurs the highest UAV energy consumption among the evaluated variants, at 15.2~kJ, confirming that the one-step action prior alone is insufficient for long-horizon energy-aware planning. Through Logit Fusion, PPO can correct energy-inefficient prior preferences while retaining their connectivity benefits. The MLP comparison further shows that the observed gain arises from both the supervised construction of the action prior and the choice of prior model.

\subsubsection{Sensitivity to the Semantic Fusion Coefficient}

\begin{figure}[ht]
    \centering
    \includegraphics[width=.8\linewidth]
    {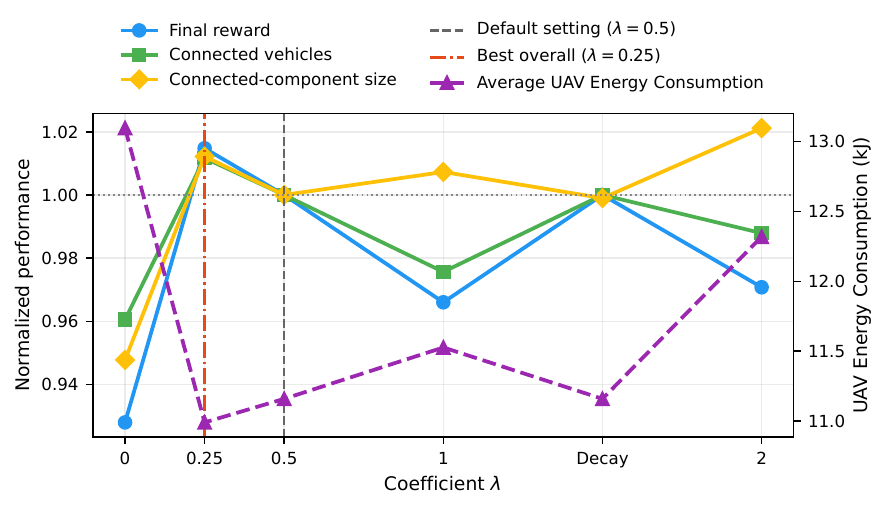}
    \caption{Sensitivity of SA-PPO to the semantic fusion coefficient $\lambda$. The reward and connectivity metrics are normalized by their values under the default setting $\lambda=0.5$, while the UAV energy consumption is reported in absolute values. ``Decay'' denotes a schedule that progressively reduces $\lambda$ during training.}
    \label{fig:lambda_sensitivity}
\end{figure}

To examine whether SA-PPO depends on a narrowly tuned fusion coefficient, we evaluate fixed coefficients $\lambda\in\{0, \allowbreak 0.25, \allowbreak 0.5, \allowbreak 1.0, \allowbreak 2.0\}$ and an additional decaying schedule that progressively reduces the semantic fusion strength during training, while keeping the remaining settings unchanged. As shown in Fig. \ref{fig:lambda_sensitivity}, the final reward and connectivity metrics are normalized by their respective values under the default setting $\lambda=0.5$, while the UAV energy consumption is reported in absolute values.

Without semantic fusion, i.e., $\lambda=0$, the normalized final reward decreases from 1.000 to 0.928, while the normalized number of connected vehicles and connected-component size decrease from 1.000 to 0.961 and 0.948, respectively. These changes correspond to reductions of 7.19\%, 3.94\%, and 5.22\%. Meanwhile, the UAV energy consumption increases from approximately 11.2 kJ to 13.1 kJ, confirming that the semantic prior effectively reduces inefficient exploration.

A moderate fixed fusion coefficient achieves the best overall connectivity--energy tradeoff. When $\lambda=0.25$, the normalized final reward, number of connected vehicles, and connected-component size increase to approximately 1.015, 1.012, and 1.011, respectively, while the UAV energy consumption decreases to approximately 11.0 kJ. In contrast, further increasing the fixed coefficient reduces the overall reward and generally increases energy consumption. Although $\lambda=2.0$ produces the largest connected-component size, it yields a lower reward and substantially higher energy consumption.

The decaying schedule yields normalized reward and connectivity metrics close to 1.000, with energy consumption comparable to the default fixed setting $\lambda=0.5$, but does not outperform the moderate fixed setting $\lambda=0.25$. This result directly examines the possibility of assigning different levels of trust to the LLM prior during different training stages and shows that a simple time-decaying schedule is not necessarily superior. Together, the fixed and decaying settings show that the benefit of Logit Fusion does not depend on a narrowly selected coefficient.

The semantic-logit cache substantially reduces the cost of repeated LLM inference because the discrete environment representation leads to frequent state recurrence within and across training runs. Without caching, one SA-PPO training run requires approximately 2 hours on an RTX 4080 Super GPU, whereas an initially empty cache reduces the first run to about 7 minutes. When the cache is shared across the $\lambda$-sensitivity experiments, including the decaying-coefficient setting, the second run encounters only one cache miss and all subsequent runs achieve a 100\% hit rate, reducing the training time to approximately 1.5 minutes per run. Since $\lambda$ affects only policy-side fusion and does not change the LLM outputs, this reuse introduces no approximation and does not alter the experimental results.

\subsubsection{Robustness and Generalization}

\begin{table}[ht]
  \centering
  \caption{Temporal robustness under different task start times. The best result at each start time is shown in bold.}
  \label{tab:temporal_robustness}
  \setlength{\tabcolsep}{4.5pt}
  \renewcommand{\arraystretch}{1.08}
  \begin{tabular}{lrrrrrr}
    \toprule
    \textbf{Method} & \textbf{08:00} & \textbf{10:00} & \textbf{12:00} & \textbf{14:00} & \textbf{16:00} & \textbf{18:00} \\
    \midrule
    \multicolumn{7}{c}{\textit{Average number of vehicles in connected components $\uparrow$}} \\
    \midrule
    Soft Actor-Critic & 36.1 & 59.3 & 44.9 & 76.3 & 82.1 & 60.7 \\
    Vanilla PPO & 38.4 & 58.2 & 44.0 & 74.7 & \textbf{86.2} & 62.2 \\
    GAT-PPO & 37.1 & 58.6 & 44.5 & 75.1 & 78.8 & 61.4 \\
    SA-PPO & \textbf{40.4} & \textbf{59.9} & \textbf{51.8} & \textbf{77.4} & 84.5 & \textbf{63.9} \\
    \midrule
    \multicolumn{7}{c}{\textit{Average connected-component size $\uparrow$}} \\
    \midrule
    Soft Actor-Critic & 16.7 & 20.3 & 18.1 & 21.7 & 18.8 & 15.7 \\
    Vanilla PPO & 18.7 & 20.8 & 18.4 & 22.3 & 20.3 & 17.2 \\
    GAT-PPO & 18.1 & 21.1 & 18.5 & 22.9 & 19.3 & 17.1 \\
    SA-PPO & \textbf{20.3} & \textbf{21.6} & \textbf{22.2} & \textbf{23.2} & \textbf{20.6} & \textbf{17.6} \\
    \midrule
    \multicolumn{7}{c}{\textit{Average UAV energy consumption (kJ) $\downarrow$}} \\
    \midrule
    Soft Actor-Critic & \textbf{10.7} & \textbf{10.7} & \textbf{10.7} & \textbf{10.7} & \textbf{10.7} & \textbf{10.7} \\
    Vanilla PPO & 14.3 & 14.3 & 14.4 & 15.3 & 15.0 & 15.0 \\
    GAT-PPO & 15.8 & 16.1 & 16.9 & 15.4 & 16.8 & 16.1 \\
    SA-PPO & 10.9 & 11.2 & 11.0 & 11.7 & 10.9 & 11.5 \\
    \bottomrule
  \end{tabular}
\end{table}

To evaluate the robustness of the algorithms against temporal distribution shifts in traffic patterns, all models are trained using traffic-flow data starting at 08:20 and tested over six distinct periods from 08:00 to 18:00 with identical durations. As shown in Table \ref{tab:temporal_robustness}, SA-PPO achieves the highest average number of vehicles in connected components in five of the six periods and the largest average connected-component size in all six periods. In the sparse 12:00 scenario, SA-PPO increases the two connectivity metrics from 44.0 and 18.4 for Vanilla PPO to 51.8 and 22.2, corresponding to improvements of 17.7\% and 20.7\%, respectively. In the dense 16:00 scenario, although its average number of vehicles in connected components is slightly lower than that of Vanilla PPO (84.5 versus 86.2), SA-PPO achieves a larger average connected-component size (20.6 versus 20.3) while reducing the average UAV energy consumption from 15.0~kJ to 10.9~kJ per time slot, corresponding to a reduction of 27.3\%. These results demonstrate that SA-PPO maintains a favorable connectivity--energy trade-off under unseen temporal traffic distributions on the same road topology.

\begin{figure}[ht]
    \centering
    \includegraphics[width=.8\linewidth]
    {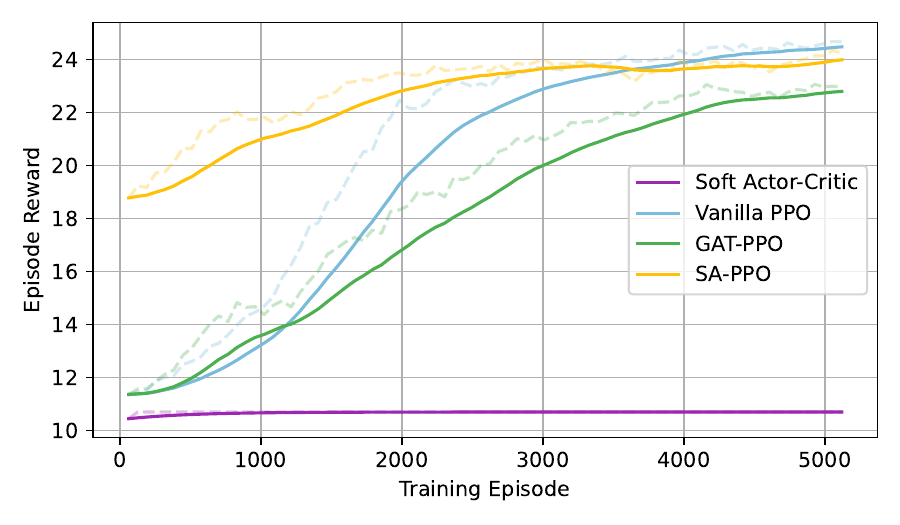}
    \caption{Training performance on the Jinan road network.}
    \label{fig:jinan_train_reward}
\end{figure}

To further evaluate the applicability of the proposed framework across different urban topologies, we additionally conduct experiments on a road network from Jinan. The Jinan RTG contains 50 candidate intersections and uses a different road structure and RSU configuration from the 47-node, 88-edge Shenzhen environment. Owing to the different topology and action-space dimension, the SFT data, semantic model, and DRL agents are independently reconstructed and trained for the Jinan environment. This is a cross-city re-instantiation test rather than zero-shot transfer. As shown in Fig. \ref{fig:jinan_train_reward}, SA-PPO reaches the high-reward region substantially earlier than the baseline algorithms and maintains the highest reward during most of the training episodes. Although Vanilla PPO eventually achieves a slightly higher asymptotic reward, SA-PPO remains competitive and clearly outperforms GAT-PPO and SAC. These results indicate that the proposed framework preserves its sample-efficiency advantage when independently trained on a structurally different urban road network.

\section{Conclusion}\label{sec:conclusion}

To address VANET fragmentation in the considered urban environments, we proposed the SA-DRL framework, which quantifies network fragmentation using the RTG and DCG and incorporates LLM-derived semantic priors into PPO through Logit Fusion. In the main Shenzhen comparison, SA-PPO reached the final converged reward level of Vanilla PPO using 28.6\% as many training episodes, improved the two reported connectivity metrics by 7.9\% and 8.7\%, and reduced UAV energy consumption by 21.3\%. Across six unseen traffic periods on the same topology, SA-PPO maintained a favorable connectivity--energy trade-off. When the complete framework was independently reconstructed and trained for the Jinan topology, it retained its sample-efficiency advantage, although Vanilla PPO achieved a slightly higher asymptotic reward. These results support improved sample efficiency and robustness in the evaluated settings, while broader transfer and multi-UAV operation remain outside the scope of this study. 

\bibliographystyle{IEEEtran}
\bibliography{reference}

@ARTICLE{6g_auto_driving,
  author={Nguyen, Van-Linh and Hwang, Ren-Hung and Lin, Po-Ching and Vyas, Abhishek and Nguyen, Van-Tao},
  journal={IEEE Network}, 
  title={Toward the Age of Intelligent Vehicular Networks for Connected and Autonomous Vehicles in {6G}}, 
  year={2023},
  volume={37},
  number={3},
  pages={44-51},
  doi={10.1109/MNET.010.2100509}}

@ARTICLE{vanet_survey,
  author={Mahi, Md. Julkar Nayeen and Chaki, Sudipto and Ahmed, Shamim and Biswas, Milon and Kaiser, M. Shamim and Islam, Mohammad Shahidul and Sookhak, Mehdi and Barros, Alistair and Whaiduzzaman, Md},
  journal={IEEE Access}, 
  title={A Review on {VANET} Research: Perspective of Recent Emerging Technologies}, 
  year={2022},
  volume={10},
  number={},
  pages={65760-65783},
  doi={10.1109/ACCESS.2022.3183605}}

@article{shadow_fading,
  title={A measurement based shadow fading model for vehicle-to-vehicle network simulations},
  author={Abbas, Taimoor and Sj{\"o}berg, Katrin and Karedal, Johan and Tufvesson, Fredrik},
  journal={Int. J. Antennas Propag.},
  volume={2015},
  number={1},
  pages={190607},
  year={2015},
  publisher={Wiley Online Library}}

@ARTICLE{mobility_channel,
  author={Akhtar, Nabeel and Ergen, Sinem Coleri and Ozkasap, Oznur},
  journal={IEEE Trans. Veh. Technol.}, 
  title={Vehicle Mobility and Communication Channel Models for Realistic and Efficient Highway VANET Simulation}, 
  year={2015},
  volume={64},
  number={1},
  pages={248-262},
  doi={10.1109/TVT.2014.2319107}}

@ARTICLE{fragmented_subnets,
  author={Oubbati, Omar Sami and Atiquzzaman, Mohammed and Baz, Abdullah and Alhakami, Hosam and Ben-Othman, Jalel},
  journal={IEEE Trans. Veh. Technol.}, 
  title={Dispatch of {UAV}s for Urban Vehicular Networks: A Deep Reinforcement Learning Approach}, 
  year={2021},
  volume={70},
  number={12},
  pages={13174-13189},
  doi={10.1109/TVT.2021.3119070}}

@article{rsu_cost,
  title={Efficient deployment of roadside units in vehicular networks using optimization methods},
  author={Lehsaini, Mohamed and Gaouar, Nihal and Nebbou, Tawfiq},
  journal={Int. J. Commun. Syst.},
  volume={35},
  number={14},
  pages={e5265},
  year={2022},
  publisher={Wiley Online Library}}

@ARTICLE{rsu_residence,
  author={Clancy, Joseph and Mullins, Darragh and Deegan, Brian and Horgan, Jonathan and Ward, Enda and Eising, Ciarán and Denny, Patrick and Jones, Edward and Glavin, Martin},
  journal={IEEE Commun. Surv. Tutorials}, 
  title={Wireless Access for {V2X} Communications: Research, Challenges and Opportunities}, 
  year={2024},
  volume={26},
  number={3},
  pages={2082-2119},
  doi={10.1109/COMST.2024.3384132}}

@ARTICLE{uav_los,
  author={Gapeyenko, Margarita and Moltchanov, Dmitri and Andreev, Sergey and Heath, Robert W.},
  journal={IEEE Trans. Wireless Commun.}, 
  title={Line-of-Sight Probability for mmWave-Based {UAV} Communications in {3D} Urban Grid Deployments}, 
  year={2021},
  volume={20},
  number={10},
  pages={6566-6579},
  doi={10.1109/TWC.2021.3075099}}

@ARTICLE{uav_survey,
  author={Kurunathan, Harrison and Huang, Hailong and Li, Kai and Ni, Wei and Hossain, Ekram},
  journal={IEEE Commun. Surv. Tutorials}, 
  title={Machine Learning-Aided Operations and Communications of Unmanned Aerial Vehicles: A Contemporary Survey}, 
  year={2024},
  volume={26},
  number={1},
  pages={496-533},
  doi={10.1109/COMST.2023.3312221}}

@ARTICLE{np_hard,
  author={Sabzehali, Javad and Shah, Vijay K. and Fan, Qiang and Choudhury, Biplav and Liu, Lingjia and Reed, Jeffrey H.},
  journal={IEEE Internet Things J.}, 
  title={Optimizing Number, Placement, and Backhaul Connectivity of Multi-{UAV} Networks}, 
  year={2022},
  volume={9},
  number={21},
  pages={21548-21560},
  doi={10.1109/JIOT.2022.3184323}}

@article{heuristic_survey,
  title={Path planning optimization in unmanned aerial vehicles using meta-heuristic algorithms: A systematic review},
  author={Yahia, Hazha Saeed and Mohammed, Amin Salih},
  journal={Environ. Monit. Assess.},
  volume={195},
  number={1},
  pages={30},
  year={2023},
  publisher={Springer}}

@ARTICLE{drl_uav,
  author={Bai, Yu and Zhao, Hui and Zhang, Xin and Chang, Zheng and Jäntti, Riku and Yang, Kun},
  journal={IEEE Commun. Surv. Tutorials}, 
  title={Toward Autonomous Multi-{UAV} Wireless Network: A Survey of Reinforcement Learning-Based Approaches}, 
  year={2023},
  volume={25},
  number={4},
  pages={3038-3067},
  doi={10.1109/COMST.2023.3323344}}

@article{ppo,
  title={Proximal policy optimization algorithms},
  author={Schulman, John and Wolski, Filip and Dhariwal, Prafulla and Radford, Alec and Klimov, Oleg},
  journal={arXiv preprint arXiv:1707.06347},
  year={2017}
}

@ARTICLE{ppo_uav,
  author={Guan, Yue and Zou, Sai and Peng, Haixia and Ni, Wei and Sun, Yanglong and Gao, Hongfeng},
  journal={IEEE Internet Things J.}, 
  title={Cooperative {UAV} Trajectory Design for Disaster Area Emergency Communications: A Multiagent {PPO} Method}, 
  year={2024},
  volume={11},
  number={5},
  pages={8848-8859},
  doi={10.1109/JIOT.2023.3320796}}

@ARTICLE{llm_uav,
  author={Javaid, Shumaila and Fahim, Hamza and He, Bin and Saeed, Nasir},
  journal={IEEE Open J. Veh. Technol.}, 
  title={Large Language Models for {UAV}s: Current State and Pathways to the Future}, 
  year={2024},
  volume={5},
  number={},
  pages={1166-1192},
  doi={10.1109/OJVT.2024.3446799}}

@article{llm_uav_1,
  title={Next-Generation LLM for UAV: From Natural Language to Autonomous Flight},
  author={Yuan, Liangqi and Deng, Chuhao and Han, Dong-Jun and Hwang, Inseok and Brunswicker, Sabine and Brinton, Christopher G},
  journal={arXiv preprint arXiv:2510.21739},
  year={2025}
}

@article{gpt4,
  title={Sparks of artificial general intelligence: Early experiments with gpt-4},
  author={Bubeck, S{\'e}bastien and Chandrasekaran, Varun and Eldan, Ronen and Gehrke, Johannes and Horvitz, Eric and Kamar, Ece and Lee, Peter and Lee, Yin Tat and Li, Yuanzhi and Lundberg, Scott and others},
  journal={arXiv preprint arXiv:2303.12712},
  year={2023}}

@INPROCEEDINGS{llm_navigation,
  author={Zu, Weiqin and Song, Wenbin and Chen, Ruiqing and Guo, Ze and Sun, Fanglei and Tian, Zheng and Pan, Wei and Wang, Jun},
  booktitle={2024 IEEE Int. Conf. Robot. Autom. (ICRA)}, 
  title={Language and Sketching: An {LLM}-driven Interactive Multimodal Multitask Robot Navigation Framework}, 
  year={2024},
  volume={},
  number={},
  pages={1019-1025},
  doi={10.1109/ICRA57147.2024.10611462}}

@ARTICLE{rel1,
  author={Mokhtari, Somayeh and Nouri, Nima and Abouei, Jamshid and Avokh, Avid and Plataniotis, Konstantinos N.},
  journal={IEEE Internet Things J.}, 
  title={Relaying Data With Joint Optimization of Energy and Delay in Cluster-Based UAV-Assisted {VANET}s}, 
  year={2022},
  volume={9},
  number={23},
  pages={24541-24559},
  doi={10.1109/JIOT.2022.3188563}}

@ARTICLE{rel2,
  author={Chughtai, Omer and Qadri, Nadia Nawaz and Kaleem, Zeeshan and Yuen, Chau},
  journal={IEEE Access}, 
  title={Drone-Assisted Cooperative Routing Scheme for Seamless Connectivity in {V2X} Communication}, 
  year={2024},
  volume={12},
  number={},
  pages={17369-17381},
  doi={10.1109/ACCESS.2024.3359273}}

@ARTICLE{rel3,
  author={Andreou, Andreas and Mavromoustakis, Constandinos X. and Batalla, Jordi Mongay and Markakis, Evangelos K. and Mastorakis, George},
  journal={IEEE Trans. Intell. Transp. Syst.}, 
  title={UAV-Assisted {RSU}s for {V2X} Connectivity Using Voronoi Diagrams in {6G}+ Infrastructures}, 
  year={2023},
  volume={24},
  number={12},
  pages={15855-15865},
  doi={10.1109/TITS.2023.3273716}}

@ARTICLE{rel4,
  author={Fan, Xiying and Zhang, Haijun and Huang, Yao and Su, Yu and Li, Haojin and Huo, Jiahao and Sun, Chen and Hao, Sheng and Zhen, Li},
  journal={IEEE Trans. Veh. Technol.}, 
  title={Temporal Data Dissemination in {UAV}-Assisted {VANET}s Through Time-Varying Graphs}, 
  year={2024},
  volume={73},
  number={10},
  pages={14835-14846},
  doi={10.1109/TVT.2024.3401230}}

@ARTICLE{rel5,
  author={Wei, Xianglin and Cai, Lingfeng and Wei, Nan and Zou, Peng and Zhang, Jin and Subramaniam, Suresh},
  journal={IEEE Internet Things J.}, 
  title={Joint {UAV} Trajectory Planning, {DAG} Task Scheduling, and Service Function Deployment Based on {DRL} in UAV-Empowered Edge Computing}, 
  year={2023},
  volume={10},
  number={14},
  pages={12826-12838},
  doi={10.1109/JIOT.2023.3257291}}

@ARTICLE{rel6,
  author={Yao, Yuanyuan and Lv, Ke and Huang, Sai and Xiang, Wei},
  journal={IEEE Trans. Veh. Technol.}, 
  title={{3D} Deployment and Energy Efficiency Optimization Based on {DRL} for {RIS}-Assisted Air-to-Ground Communications Networks}, 
  year={2024},
  volume={73},
  number={10},
  pages={14988-15003},
  doi={10.1109/TVT.2024.3405608}}

@ARTICLE{rel7,
  author={Chen, Jingxuan and Huang, Dianrun and Wang, Yijie and Yu, Ziping and Zhao, Zhongliang and Cao, Xianbin and Liu, Yang and Quek, Tony Q. S. and Oliver Wu, Dapeng},
  journal={IEEE Trans. Mach. Learn. Commun. Netw.}, 
  title={Enhancing Routing Performance Through Trajectory Planning With {DRL} in {UAV}-Aided {VANET}s}, 
  year={2025},
  volume={3},
  number={},
  pages={517-533},
  doi={10.1109/TMLCN.2025.3558204}}

@ARTICLE{rel8,
  author={Hou, Peng and Huang, Yi and Zhu, Hongbin and Lu, Zhihui and Huang, Shin-Chia and Yang, Yang and Chai, Hongfeng},
  journal={IEEE Internet Things J.}, 
  title={Distributed {DRL}-Based Intelligent Over-the-Air Computation in Unmanned Aerial Vehicle Swarm-Assisted Intelligent Transportation System}, 
  year={2024},
  volume={11},
  number={21},
  pages={34382-34397},
  doi={10.1109/JIOT.2024.3418882}}

@article{rel9,
  title={{UAV} Visual Path Planning Using Large Language Models},
  author={Samma, Hussein and El-Ferik, Sami},
  journal={Transp. Res. Procedia},
  volume={84},
  pages={339--345},
  year={2025},
  publisher={Elsevier}}

@article{rel10,
  title={{LLM}-Land: Large Language Models for Context-Aware Drone Landing},
  author={Cai, Siwei and Wu, Yuwei and Zhou, Lifeng},
  journal={arXiv preprint arXiv:2505.06399},
  year={2025}}

@ARTICLE{rel11,
  author={Zhou, Qian and Wu, Jiayang and Zhu, Mengyue and Zhou, Yuhang and Xiao, Fu and Zhang, Yanchun},
  journal={IEEE Trans. Knowl. Data Eng.}, 
  title={{LLM}-{QL}: A {LLM}-Enhanced Q-Learning Approach for Scheduling Multiple Parallel Drones}, 
  year={2025},
  volume={37},
  number={9},
  pages={5393-5406},
  doi={10.1109/TKDE.2025.3579386}}

@ARTICLE{rel12,
  author={Wu, Jiahao and You, Hengxu and Sun, Bowen and Du, Jing},
  journal={IEEE Access}, 
  title={{LLM}-Driven Pareto-Optimal Multi-Mode Reinforcement Learning for Adaptive {UAV} Navigation in Urban Wind Environments}, 
  year={2025},
  volume={13},
  number={},
  pages={163550-163570},
  doi={10.1109/ACCESS.2025.3611336}}

@article{idm,
  title={Limitations and improvements of the intelligent driver model ({IDM})},
  author={Albeaik, Saleh and Bayen, Alexandre and Chiri, Maria Teresa and Gong, Xiaoqian and Hayat, Amaury and Kardous, Nicolas and Keimer, Alexander and McQuade, Sean T and Piccoli, Benedetto and You, Yiling},
  journal={SIAM J. Appl. Dyn. Syst.},
  volume={21},
  number={3},
  pages={1862--1892},
  year={2022},
  publisher={SIAM}}

@ARTICLE{los,
  author={Al-Hourani, Akram and Kandeepan, Sithamparanathan and Lardner, Simon},
  journal={IEEE Wireless Commun. Lett.}, 
  title={Optimal {LAP} Altitude for Maximum Coverage}, 
  year={2014},
  volume={3},
  number={6},
  pages={569-572},
  doi={10.1109/LWC.2014.2342736}}

@ARTICLE{dtrl,
  author={Yuan, Wenke and Cao, Gaoxiang and Hou, Yunpeng and Wang, Jian and Chen, Shuangwu and He, Huasen and Yang, Jian},
  journal={IEEE Trans. Commun.}, 
  title={Deep Transfer Reinforcement Learning Based Exploration Enhanced Multi-{UAV} Trajectory Planning}, 
  year={2025},
  volume={},
  number={},
  pages={1-1},
  doi={10.1109/TCOMM.2025.3635773}}

@ARTICLE{uav_energy,
  author={Ding, Yu and Zhang, Qingqing and Lu, Weidang and Zhao, Nan and Nallanathan, Arumugam and Wang, Xianbin and Yang, Xiaoniu},
  journal={IEEE Trans. Wireless Commun.}, 
  title={Collaborative Communication and Computation for Secure {UAV}-Enabled {MEC} Against Active Aerial Eavesdropping}, 
  year={2024},
  volume={23},
  number={11},
  pages={15915-15929},
  doi={10.1109/TWC.2024.3435017}}

@article{lora,
  title={Lora: Low-rank adaptation of large language models.},
  author={Hu, Edward J and Shen, Yelong and Wallis, Phillip and Allen-Zhu, Zeyuan and Li, Yuanzhi and Wang, Shean and Wang, Lu and Chen, Weizhu and others},
  journal={ICLR},
  volume={1},
  number={2},
  pages={3},
  year={2022}}

@Article{dataset,
  author={Yu, Fudan and Yan, Huan and Chen, Rui and Zhang, Guozhen and Liu, Yu and Chen, Meng and Li, Yong},
  title={City-scale Vehicle Trajectory Data from Traffic Camera Videos},
  journal={Sci. Data},
  year={2023},
  month={Oct},
  day={17},
  volume={10},
  number={1},
  pages={711},
  issn={2052-4463},
  doi={10.1038/s41597-023-02589-y},
  url={https://doi.org/10.1038/s41597-023-02589-y}}

@inproceedings{vllm,
  title={Efficient Memory Management for Large Language Model Serving with PagedAttention},
  author={Woosuk Kwon and Zhuohan Li and Siyuan Zhuang and Ying Sheng and Lianmin Zheng and Cody Hao Yu and Joseph E. Gonzalez and Hao Zhang and Ion Stoica},
  booktitle={Proc. 29th ACM Symp. Oper. Syst. Princ. (SOSP)},
  year={2023}}

@inproceedings{llama_factory,
  title={LlamaFactory: Unified Efficient Fine-Tuning of 100+ Language Models},
  author={Yaowei Zheng and Richong Zhang and Junhao Zhang and Yanhan Ye and Zheyan Luo and Zhangchi Feng and Yongqiang Ma},
  booktitle={Proc. 62nd Annu. Meeting Assoc. Comput. Ling. (ACL)},
  address={Bangkok, Thailand},
  publisher={Association for Computational Linguistics},
  year={2024},
  url={http://arxiv.org/abs/2403.13372}}

@ARTICLE{cmp_algo,
  author={Goudarzi, Shidrokh and Ahmad Soleymani, Seyed and Hossein Anisi, Mohammad and Jindal, Anish and Xiao, Pei},
  journal={IEEE Internet Things J.}, 
  title={Optimizing {UAV}-Assisted Vehicular Edge Computing With Age of Information: An {SAC}-Based Solution}, 
  year={2025},
  volume={12},
  number={5},
  pages={4555-4569},
  doi={10.1109/JIOT.2025.3529836}}

@INPROCEEDINGS{hover_parameter,
  author={Qiu, Chenrui and De Simone, Lorenzo and Zhu, Yongxu and Tham, Mau-Luen and Dagiuklas, Tasos},
  booktitle={ICC 2025 - IEEE International Conference on Communications}, 
  title={Coverage Optimization Approach in Aerial-Ground Integrated Wireless Networks}, 
  year={2025},
  volume={},
  number={},
  pages={5011-5016},
  doi={10.1109/ICC52391.2025.11161611}}

@Article{communication_parameter,
AUTHOR = {Yang, Biao and Xiong, Xuanrui and Liu, He and Jia, Yumei and Gao, Yunli and Tolba, Amr and Zhang, Xingguo},
TITLE = {Unmanned Aerial Vehicle Assisted Post-Disaster Communication Coverage Optimization Based on Internet of Things Big Data Analysis},
JOURNAL = {Sensors},
VOLUME = {23},
YEAR = {2023},
NUMBER = {15},
ARTICLE-NUMBER = {6795},
URL = {https://www.mdpi.com/1424-8220/23/15/6795},
PubMedID = {37571578},
ISSN = {1424-8220},
DOI = {10.3390/s23156795}
}

@ARTICLE{propulsion_parameter,
  author={Mamaghani, Milad Tatar and Hong, Yi},
  journal={IEEE Transactions on Vehicular Technology}, 
  title={Terahertz Meets Untrusted UAV-Relaying: Minimum Secrecy Energy Efficiency Maximization via Trajectory and Communication Co-Design}, 
  year={2022},
  volume={71},
  number={5},
  pages={4991-5006},
  doi={10.1109/TVT.2022.3150011}}

\end{document}